%% file: main.tex
\documentclass[10pt,twocolumn,letterpaper]{article}

\usepackage[pagenumbers]{wacv}

\definecolor{wacvblue}{rgb}{0.21,0.49,0.74}

\usepackage[pagebackref,breaklinks,colorlinks,allcolors=wacvblue]{hyperref}
\hypersetup{
    pdftitle={Do Pretrained Backbones Transfer to Physical Sensors? A Systematic Study of Multimodal Initialization Across the Sensor Stack},
    pdfauthor={Fabio Montello, Ronja Güldenring, Lucas Mogensen Sandby, Christoffer Koo Øhrstrøm, Lazaros Nalpantidis},
    pdfsubject={IEEE/CVF Winter Conference on Applications of Computer Vision (WACV) 2027},
    pdfkeywords={multimodal pretraining, masked autoencoders, sensor fusion, autonomous driving, thermal imaging, event cameras}
}
\usepackage{graphicx}
\usepackage{booktabs}
\usepackage{tabu}
\usepackage{makecell}
\usepackage{multirow}
\usepackage{array}
\usepackage{stfloats}
\usepackage{amsmath}
\usepackage{xfrac}
\usepackage{placeins}
\usepackage[skip=6pt]{caption}
\usepackage{enumitem}

\def\confName{WACV}
\def\confYear{2027}

\title{Do Pretrained Backbones Transfer to Physical Sensors?\\A Systematic Study of Multimodal Initialization Across the Sensor Stack}

\author{
    Fabio Montello \quad
    Ronja Güldenring \quad
    Lucas Mogensen Sandby \\
    Christoffer Koo Øhrstrøm \quad
    Lazaros Nalpantidis \\
    Technical University of Denmark \\
    {\tt\small \{fabmo, lanalpa\}@dtu.dk}
}

\begin{document}
\maketitle

\begin{abstract}
Transfer learning from pretrained backbones has become standard practice in multi-sensor computer vision---yet two questions remain open: which pretrained weights transfer best to multi-sensor setups, and can in-domain pretraining outperform off-the-shelf solutions? Frameworks like MultiMAE demonstrate the utility of masked autoencoding in a multi-modal context, but rely entirely on dense, model-derived modalities. It is unclear how robustly these methods translate to real-world applications that use physical sensors, where the aggregation of diverse public datasets for large-scale pretraining inherently introduces modality sparsity. To investigate this systematically, we curate a multi-sensor autonomous driving collection of $2.82M$ images from 16 publicly available datasets, incorporating both physical hardware modalities (RGB, LWIR, NIR, Event) and model-derived modalities (Stereo Depth, Segmentation). We use this collection for in-domain pretraining based on the MultiMAE approach. We adapt MultiMAE's loss formulation to accommodate sparse samples, and study the effect of an enforced encoder bottleneck for modality alignment and of additional losses specific to dense reconstruction (SSIM and FFT). We then systematically compare against common pretrained backbones (ViT, MAE, DINO and MultiMAE) across downstream sensor fusion tasks. We find that MultiMAE is a surprisingly strong out-of-the-box baseline, transferring effectively even to modalities it has never seen, while our targeted modifications yield incremental but consistent improvements on multi-sensor transfer and overall robustness.
\end{abstract}

\begin{figure}[t]
    \centering
    \includegraphics[width=0.95\linewidth]{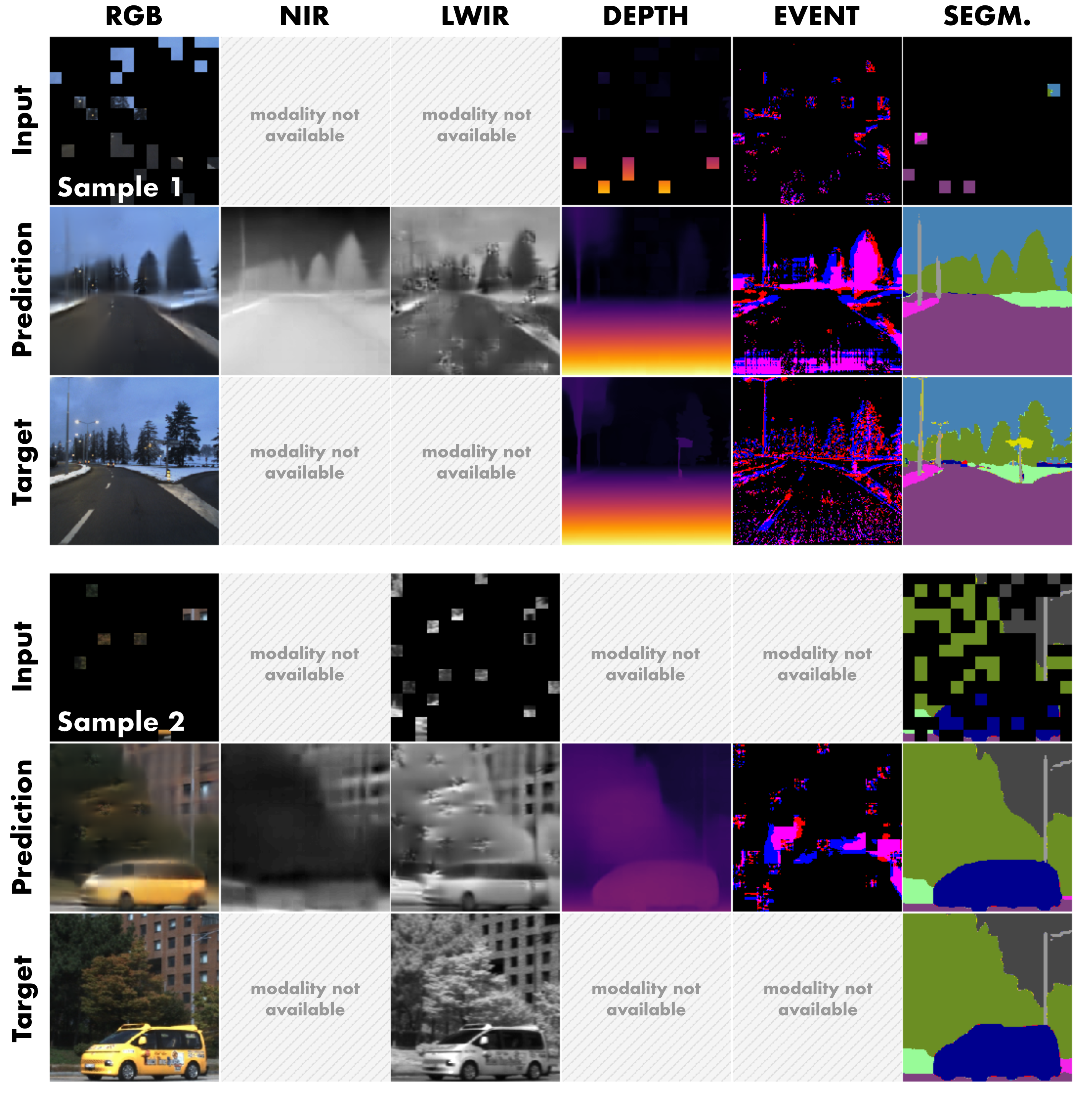}
    \caption{\textbf{Multimodal pretraining under sparse availability of sensor modalities.} Reconstruction examples from our multi-sensor pretraining pipeline. The rows display the heavily masked sparse inputs (top), the dense model predictions (middle), and the ground-truth targets (bottom) across physically collected hardware sensors (RGB, NIR, LWIR, Event) and model-derived modalities (Depth, Segmentation). Empty panels denote missing modalities due to dataset sparsity (no sample contains all modalities).}
    \label{fig:samples}
\end{figure}

\section{Introduction}
\label{sec:intro}

When finetuning for specific computer vision tasks, transfer learning from large-scale pretrained backbones is a de facto standard approach in modern computer vision. Among many different pretraining methods, masked autoencoding (MAE) \cite{he2022masked} has emerged as highly effective for self-supervised representation learning, where the model learns spatial and semantic features through the reconstruction task of heavily masked portions of an image. Frameworks such as MultiMAE \cite{multimae, 2023_4m, 2024_4m21} have expanded the idea of masked autoencoding to a multi-modal context, demonstrating that the embedding representation can further benefit from the joint reconstruction of multiple domains. This approach relies on additional generated modalities derived from RGB images, i.e. pseudo-depth and semantic maps.

\vspace{2pt} \noindent However, when finetuning for diverse real-sensor modalities, like LWIR/NIR images or Event streams, it is not obvious whether the representations learned from dense model-derived domains will transfer effectively or fall short compared to weights pretrained on physically collected data. We therefore ask: across common pretrained backbones, \emph{which weights transfer best to physical sensor finetuning?} And \emph{can a tailored in-domain solution outperform them?}
To investigate this systematically, we curate what is, to the best of our knowledge, the widest autonomous driving (AD) dataset collection in terms of range of sensors, modalities and weather scenarios, composed of 16 publicly available datasets, containing approximately $800K$ samples and $2.82M$ images — here, a \textit{sample} denotes one aligned scene instance, while a \textit{unique image} denotes one modality-specific frame within that sample (e.g. a sample with RGB, Depth, and Segmentation contributes three unique images). This collection explicitly incorporates both physical hardware modalities---RGB, Long-Wave Infrared (LWIR), Near Infrared (NIR) and Event streams---as well as model-derived modalities---relative Stereo Depth and Segmentation maps. \\

Achieving the data scale necessary for pretraining makes custom data collection expensive; aggregating publicly available datasets, while practical, inherently introduces sparsity: one dataset might provide RGB and NIR pairs, while another captures RGB alongside Event streams. Existing multimodal pretraining pipelines need to be adapted to account for this induced heterogeneity.
Our collection is then used for in-domain pretraining investigation of the adaptability of the MultiMAE approach to the physical sensor data in a sparse context. We investigate three batch reduction strategies under sparse sampling conditions to prevent gradient domination of highly represented modalities (e.g. RGB). We also test the effect of an injected bottleneck between the encoder and the decoders, as a substitute for cross-attention. We do this with the aim of encouraging modality-agnostic representations at the encoder level, hypothesizing that this could help downstream features remain more consistent regardless of which sensor subset is available at inference time. Furthermore, we explore the contribution of additional objectives—such as Structural Similarity (SSIM) and Fast Fourier Transform (FFT) losses. While these losses are beneficial for structured image reconstruction in general \cite{kraus2024masked}, they intuitively offer even greater advantages for physical sensors, where samples are rich in fine visual details.

\vspace{2pt} \noindent We evaluate these pretrained weights on the tasks of semantic segmentation and object detection. We find that, out of all the tested pretraining weights, standard MultiMAE serves as a remarkably strong out-of-the-box baseline, adapting surprisingly well to new multimodal contexts. Nevertheless, our targeted modifications yield incremental but consistent accuracy gains in most settings and improved overall robustness across downstream benchmarks.
We summarize our main contributions as follows:
\begin{itemize}[topsep=2pt,itemsep=1pt,leftmargin=*]
    \item We extensively \textit{evaluate pretrained weight initializations} for multi-sensor downstream finetuning---comparing ViT, MAE, DINO, MultiMAE, and our in-domain variants. We demonstrate that while off-the-shelf MultiMAE serves as a surprisingly robust baseline for unseen sensors, in-domain pretraining yields largely incremental gains specifically when non-RGB modalities are involved.
    \item We detail a comprehensive methodology for \textit{curating a large-scale multi-sensor pretraining collection} ($800K$ samples comprising $2.82M$ images from 4 hardware sensors and 2 model-derived modalities) as a realistic testbed for studying physical sensor transfer.
    \item We \textit{adapt standard MultiMAE pretraining for physical sensors} by testing MultiMAE's loss formulations for sparse samples, evaluating encoder bottlenecks for modality alignment, and integrating dense reconstruction losses (SSIM, FFT).
    \item \textit{We release our multi-sensor pretrained weights} in \textit{Base}, \textit{Small}, and \textit{Tiny} sizes, to also enable deployable sensor-fusion vision on compute-constrained robotics and autonomous driving platforms. Our benchmarking tool and the script to reproduce the dataset are publicly available at \href{https://github.com/DTU-PAS/sensormae}{\textit{https://github.com/DTU-PAS/sensormae}}.
\end{itemize}

\section{Related Work}
\label{sec:related_work}

\textbf{Representation Learning.}
Pre-trained weights are a core component in modern computer vision. Self-supervised learning allows networks to learn visual structures from unlabeled data by solving specific pretext tasks. This has led to a wide variety of different training paradigms. Contrastive methods like SimCLR~\cite{SIMCLR} and MoCo~\cite{he2019momentum} force models to match different views of the same image. Self-distillation frameworks like DINO~\cite{dino} use a teacher-student setup to capture semantic properties. Masked Image Modeling has also become highly popular due to its simple setup and high efficiency~\cite{he2022masked}. The strategy is to mask random patches of an input image and train the network to reconstruct the missing pixels, forcing the understanding of spatial structures. All these methods create a strong base for downstream tasks by learning from massive, unlabeled datasets.

\vspace{2pt} \noindent \textbf{Multi-Modal Masked Autoencoders.}
MultiMAE~\cite{multimae} showed that masked autoencoding works well beyond single RGB images. It takes as input triplets of RGB, model-derived Depth, and semantic maps. The architecture uses separate tokenizers for each input stream and then drops a large portion of these tokens. A shared encoder processes only the remaining visible pieces, whereas separate decoders then attempt to reconstruct the dropped regions back together. The 4M framework~\cite{2023_4m} and its follow-up 4M-21 \cite{2024_4m21} extended this design to unstructured modalities, including text, bounding boxes and CLIP \cite{radford2021learning} features. All these solutions rely heavily on model-derived data. More recently, there have been a few studies for domain-specific applications. CA-MAE~\cite{kraus2024masked} explores specialized channel-agnostic tokenization to handle cross-channel correlations. In remote sensing, FG-MAE~\cite{wang2024feature} changes the reconstruction target from raw pixels to handcrafted features to avoid speckle noise. CMTSSL~\cite{carlesso2025curriculum} combines Masked Image Modeling with decoupled spatial-spectral jigsaw puzzles to pretrain lightweight encoders using a data-driven difficulty curriculum. These works still rely on a limited number of input modalities and on a complete dataset. In contrast, we train our framework on 6 modalities across 16 datasets, studying the effect of data sparsity on pretraining.

\vspace{2pt} \noindent \textbf{Dense Reconstruction Losses.} Standard masked autoencoders usually minimize the Mean Squared Error (MSE) over the missing patches~\cite{he2022masked}. MSE is easy to optimize, but it often creates blurry predictions. The loss forces the network to output safe averages to avoid large edge penalties. Recent works experiment with the frequency domain. For instance, Pham et al.~\cite{PhamChaMAE2025} incorporate Fast Fourier Transform (FFT) objectives into their pretraining pipeline. This helps the network overcome its natural bias toward smooth, low-frequency outputs. Spectral losses are more forgiving of small spatial shifts caused by sensor calibration errors. Structural metrics can also guide the network to capture better shape and contrast details. To the best of our knowledge, no existing method has used the Structural Similarity Index (SSIM) as an explicit loss function for masked autoencoder pretraining. We explore which of these losses works best in the case of dense structured reconstruction.

\section{Multi-Sensor Pretraining Collection}
\begin{table*}[b]
    \centering
    \caption{\textbf{Our Multi-Sensor Pretraining Collection.} Overview of the 16 datasets used and the modalities included in our pretraining collection. The Table details the sample distributions across training, validation, and test splits, alongside the standardized spatial resolutions and the sensor modalities present in each dataset.}
    \label{tab:dataset_overview}
    \resizebox{\linewidth}{!}{
    \begin{tabular}{r|c|c|c|c|c|c|c|c|c|c|c|c}
        \toprule
        \textbf{Dataset} & \multicolumn{3}{c|}{\textbf{Samples Split}} & \textbf{Resolution} & \multicolumn{6}{c|}{\textbf{Modalities}} & \textbf{Total} & \textbf{Unique} \\
         & Training & Val & Test & & RGB & Depth & Event & LWIR & NIR & Segmentation & Samples & Samples \\
        \midrule
        BDD100k \cite{Yu2018BDD100KAD} & 140,000 & 20,000 & 40,000 & $640 \times 720$ & \checkmark & & & & & \checkmark & 400,000 & 200,000 \\
        Cityscapes \cite{Cordts2016Cityscapes}& 91,716 & 2,000 & 6,100 & $640 \times 775$ & \checkmark & \checkmark & & & & \checkmark & 299,448 & 99,816 \\
        DSEC \cite{gehrig2021dsec, Gehrig3dv2021} & 10,350 & 211 & 2,245 & $600 \times 430$ & \checkmark & \checkmark & \checkmark & & & \checkmark & 51,224 & 12,806 \\
        ECMD \cite{10342726}& 52,424 & 1,036 & 0 & $640 \times 480$ & \checkmark & \checkmark & \checkmark & \checkmark & & \checkmark & 267,300 & 53,460 \\
        FMB \cite{liu2023segmif}& 1,202 & 18 & 280 & $800 \times 600$ & \checkmark & & & \checkmark & & \checkmark & 4,500 & 1,500 \\
        FR-T \cite{vertens2020heatnet}& 60,699 & 1,860 & 192 & $640 \times 650$ & \checkmark & & & \checkmark & & \checkmark & 188,253 & 62,751 \\
        KAIST-MS-Pedestrian \cite{7298706}& 11,991 & 262 & 0 & $640 \times 512$ & \checkmark & & & \checkmark & & \checkmark & 36,759 & 12,253 \\
        LLVIP \cite{jia2021llvip}& 23,690 & 360 & 6,926 & $640 \times 1024$ & \checkmark & & & \checkmark & & \checkmark & 92,928 & 30,976 \\
        MFNet \cite{8206396}& 1,568 & 392 & 393 & $640 \times 480$ & \checkmark & & & \checkmark & & \checkmark & 7,059 & 2,353 \\
        MS2 \cite{shin2023deep}& 105,906 & 2,138 & 0 & $640 \times 384$ & \checkmark & \checkmark & & \checkmark & \checkmark & \checkmark & 540,220 & 108,044 \\
        NSAVP \cite{carmichael2025dataset}& 46,988 & 954 & 0 & $639 \times 511$ & \checkmark & \checkmark & \checkmark & \checkmark & & \checkmark & 239,710 & 47,942 \\
        NTU4DRadLM \cite{ZhangZhugeLiu2023ITSC}& 19,831 & 402 & 0 & $640 \times 512$ & \checkmark & & & \checkmark & & \checkmark & 60,699 & 20,233 \\
        RASMD \cite{JIN2026103872}& 51,776 & 1,047 & 0 & $640 \times 718$ & \checkmark & & & & \checkmark & \checkmark & 158,469 & 52,823 \\
        SMOD \cite{11146667}& 8,504 & 172 & 0 & $640 \times 512$ & \checkmark & & & \checkmark & & \checkmark & 26,028 & 8,676 \\
        STheReO \cite{9981857}& 56,000 & 1,152 & 0 & $640 \times 560$ & \checkmark & \checkmark & & \checkmark & & \checkmark & 228,608 & 57,152 \\
        Vivid++ \cite{9760091}& 54,771 & 1,099 & 0 & $640 \times 480$ & \checkmark & & \checkmark & \checkmark & & \checkmark & 223,480 & 55,870 \\
        \midrule
        \textbf{\textit{Total}}& \textbf{737,416} & \textbf{33,103} & \textbf{56,136} & & \textbf{826,655} & \textbf{379,220} & \textbf{170,078} & \textbf{461,210} & \textbf{160,867} & \textbf{826,655} & \textbf{2,824,685} & \textbf{826,655} \\
        \midrule
        \textbf{\textit{Ratio}}& \textbf{89.20\%} & \textbf{4.00\%} & \textbf{6.79\%} & & \textbf{100.00\%} & \textbf{45.87\%} & \textbf{20.57\%} & \textbf{55.79\%} & \textbf{19.46\%} & \textbf{100.00\%}& &  \\
        \bottomrule
    \end{tabular}
    }
\end{table*}
\label{subsec:modalities_datasets}
In the last decade, the domain of autonomous driving has received a lot of attention, and plenty of multi-modal datasets in this field focus on particular combinations of sensors for a variety of downstream tasks. Because the raw data originates from heterogeneous sensor setups, an ensemble of such datasets will be inherently sparse. Thus, we established some guidelines to create robustness and uniformity within our collection among 6 selected modalities (RGB, model-derived Stereo Depth, LWIR, NIR, Events and Segmentation masks). Our general preprocessing methodology follows these key steps:

\vspace{2pt} \noindent \textbf{Data Extraction, Alignment and Downsampling.} For datasets provided as ROS bags, we first extract the raw sensor streams into a standardized image and array format. Modalities are spatially aligned using the provided extrinsic calibration parameters to ensure spatial consistency across heterogeneous sensors. All image pairs are projected and cropped to isolate the common Field of View (FoV). In some cases a synchronization step is also necessary among sensors collecting samples at asynchronous timestamps. In those cases, frames are matched only if they lie within a 50ms window frame. In many cases the datasets already provide synchronized and correctly overlapping images and this step is skipped. In cases where entire sequences are provided, we downsample the dataset to a standardized sample rate which can vary between $3Hz$ and $5Hz$. This reduces redundant information and creates a balance between sequence-based datasets (e.g. ECMD \cite{10342726}, DSEC \cite{gehrig2021dsec, Gehrig3dv2021}) and discrete sampled frames datasets (e.g. BDD100k \cite{Yu2018BDD100KAD}, Cityscapes \cite{Cordts2016Cityscapes}).

\vspace{2pt} \noindent \textbf{Modality-Specific Processing.} We apply specific processing techniques to certain modalities. For \textit{Event} streams, we generate 2D Event frames by collecting all events in a \textit{50ms} temporal window prior to a given image timestamp. We encode them into different channels within a $\Delta t$ interval of \textit{10ms}, keeping separate channels for positive and negative events. The output is an $H\times W \times C$ map with $C = 10$, where each pixel represents the event count in a specific $\Delta t$ interval and event polarity. For LWIR/NIR images, we apply Contrast Limited Adaptive Histogram Equalization (CLAHE) \cite{109340} to enhance local contrast and edge definitions, and all samples receive a segmentation mask. Concerning \textit{Stereo Depth} maps, these are derived from rectified left-right RGB pairs using FoundationStereo \cite{wen_trepte_aribido_kautz_gallo_birchfield}. The model output is already normalized to ensure consistent scale across datasets. Further details on the depth and mask creation are provided in Appendix \ref{sec:foundationstereo}.

\vspace{2pt} \noindent \textbf{Dimensionality Reduction and Dataset Split.} High resolution images pose a challenge when it comes to training on very large batch sizes. We perform horizontal cropping on all image pairs where the original horizontal resolution exceeds \textit{800px}. Specifically, these images are divided into overlapping vertical crops with a fixed width of \textit{640px}. We kept the original height invariant to preserve the full vertical context of the driving scene, avoiding semantic truncation of tall objects, such as pedestrians or traffic lights.
To maintain evaluation integrity within the downstream tasks, we preserve the original train/validation/test splits whenever provided by the authors, ensuring that no test set data is in the pretraining corpus. For datasets lacking a predefined split, we sample and hold out 2\% of the processed frame pairs to construct a validation split, and skip the creation of a test split to fully utilize the samples for pretraining.\\

\noindent Table \ref{tab:dataset_overview} presents a comprehensive overview of the datasets used and the modalities included in our dataset collection. In total, our compilation includes 826,655 aligned frame samples, generating 2,824,685 unique images with an average of 3.42 modalities per sample. By downsampling the original high-frequency sensor streams, we ensure that all frames are highly information-dense, with minimal redundancy in training data. The RGB modality is present across all samples as the baseline modality, along with the respective Segmentation mask. More than half of the samples include an LWIR image, and nearly half contain a dense Stereo Depth map.
We favored data collections with adverse environmental conditions (e.g. fog, heavy rain) collected at different illumination scenarios to tackle challenging edge cases, where multimodality can make a difference over a single modality.
The pretraining corpus is not publicly released, as the complex licensing terms of the 16 constituent datasets prevent direct redistribution (Appendix \ref{sec:licensing}); detailed pre-processing and reproduction guidance is provided in Appendix \ref{sec:datasets}.

\section{Method}
\label{sec:methods}
\begin{figure}[t]
    \centering
    \includegraphics[width=0.84\linewidth]{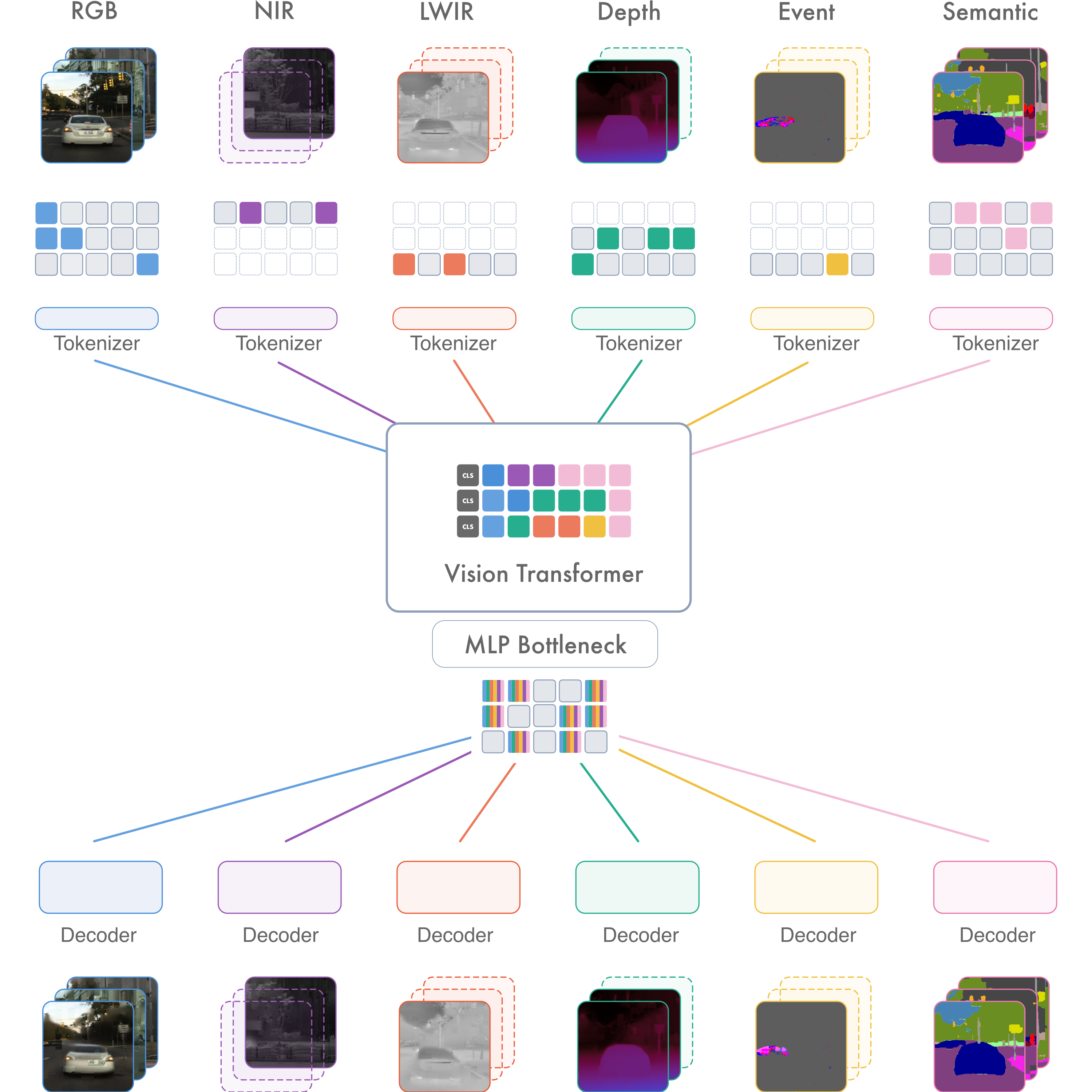}
    \caption{\textbf{Overview of our modified MultiMAE architecture designed for sparse physical sensors.} Independent tokenizers process each available sensor modality before a fixed token budget is sent to a shared ViT encoder. Instead of relying on decoder-level cross-attention, an MLP bottleneck aligns and concatenates active sensor tokens along the channel dimension (zero-filling missing ones) to produce a unified spatial representation before decoding.}
    \label{fig:main_architecture}
\end{figure}

\textit{MultiMAE} \cite{multimae} is our direct baseline and the pretraining method we extend with our dataset collection. The framework is trained on triplets of RGB, Depth and semantic images from the ImageNet1K \cite{deng2009imagenet} dataset---where the additional modalities have been derived by using respectively DPT-Hybrid \cite{ranftl2021vision} and Mask2Former \cite{cheng2022masked}. Each input domain is processed by a dedicated adapter, or tokenizer, which patchifies the incoming data and projects it into a shared embedding dimension, with the addition of a learnt positional embedding. Following the \textit{MAE} strategy, in \textit{MultiMAE} only $\sfrac{1}{6}$-th of the total tokens are kept, sampling through a Dirichlet distribution to randomly vary the proportion of kept patches per modality. \\
These sparse, unmasked tokens are concatenated into a unified sequence and processed by a shared \textit{ViT} encoder, keeping computational overhead manageable as the number of modalities grows. Domain-specific decoders then reconstruct the original inputs, each beginning with a cross-attention layer where visible patches and learned mask tokens of the target modality serve as Queries, and the full encoder output serves as Keys and Values.\\
\textit{MultiMAE} was designed for dense, complete, model-derived inputs. We adapt it for sparse, heterogeneous physical sensor data by investigating modality-balanced loss formulations, additional reconstruction objectives, and an encoder-level bottleneck for cross-modal alignment (as illustrated in Figure \ref{fig:main_architecture}). Beyond these, our base configuration differs from MultiMAE in three ways: we expand the input space to six distinct sensor modalities; we adopt a fixed token budget of 98 tokens per sample \cite{2023_4m, 2024_4m21} rather than a strict masking ratio, since variable sensor availability across samples would otherwise create uneven sequence lengths and unnecessary padding. Finally, we ensure at least one token is retained per available modality to distinguish present from absent inputs, performing the masking step before token projection to further speed up training.

\subsection{Reconstruction Losses}
The Mean Squared Error (MSE) is the standard loss approach for reconstruction tasks \cite{he2022masked, multimae, 2023_4m, 2024_4m21}. Physical sensors capture structurally rich data that MSE reconstruction tends to blur. MSE provides a direct pixel-level penalty and recovers low-frequency information effectively, but produces smoothed averages that suppress sharp edges, and ignores local spatial context entirely. We therefore combine MSE with losses that penalize structural and frequency disruptions, following the practice of \cite{kraus2024masked}.

\vspace{2pt} \noindent \textbf{SSIM Loss.}
The Structural Similarity Index Measure (SSIM) is designed to align closely with human visual perception, explicitly taking into account luminance, contrast, and structure of an image. It uses a kernel to compare local means ($\mu$), variances ($\sigma^2$), and cross-covariances ($\sigma_{PT}$) within a localized window. Given a dense prediction grid $\mathbf{P}$ and its corresponding target $\mathbf{T}$, the structural loss formulation is defined as:
\begin{equation}
L_{\text{SSIM}} = 1 - \frac{(2\mu_P\mu_T + c_1)(2\sigma_{PT} + c_2)}{(\mu_P^2 + \mu_T^2 + c_1)(\sigma_P^2 + \sigma_T^2 + c_2)}
\end{equation}
where $c_1$ and $c_2$ are constants to prevent zero division in uniform regions. SSIM has become a standard optimization metric for high-fidelity 3D scene reconstruction tasks \cite{kerbl20233d, mildenhall2021nerf}, and we therefore extend it to 2D reconstruction tasks.

\vspace{2pt} \noindent \textbf{FFT Loss.}
Following \cite{kraus2024masked}, we also test the use of a Fast Fourier Transform (FFT) loss, an alternative to SSIM loss for capturing image details.
FFT induces the optimization to align the spatial frequency spectrum, providing supervision for both macro-structures (low frequencies) and details (high frequencies). The loss projects both prediction and target into the spectral domain using a 2D real FFT ($\mathcal{F}$):
\begin{equation}
L_{\text{FFT}} = \frac{1}{|\Omega|} \left\| \log(1 + |\mathcal{F}(\mathbf{P})|) - \log(1 + |\mathcal{F}(\mathbf{T})|) \right\|_1
\end{equation}
where $\Omega$ represents the frequency grid dimensions. Amplitudes are compressed logarithmically before computing the $L_1$ distance to avoid over-representing low-frequency components.
FFT loss is forgiving of the spatial nature of the prediction: as long as the frequency component exists, it is not important where it is located spatially. This can be more tolerant of reconstruction in the context of sensor misalignment, but if used alone, FFT may produce spatial artifacts.

\subsection{Loss Balancing}
\label{sec:loss_balancing}
Aggregating datasets with different hardware setups means modality availability varies drastically across samples. In our collection, RGB is present in every sample, while Event streams appear in fewer than 21\% of the samples. Without intervention, frequently available modalities dominate training gradients and starve under-represented sensors of meaningful updates. We investigate three strategies to mitigate this imbalance ($i$ indexes samples, $j$ indexes spatial locations, throughout).

Note that the MSE loss is evaluated exclusively on the masked spatial regions, while dense complementary objectives (SSIM, FFT) are evaluated over the complete spatial grid to retain full spatial and frequency context. The total loss for any modality is therefore a composite of these, for example a weighted sum of masked MSE and dense FFT.
To formalize, let $\mathbf{P}_m$ and $\mathbf{T}_m$ denote the prediction and target tensors for a given modality $m$, both of shape $(N_m, K_m)$, where $N_m$ is the number of samples in the batch containing modality $m$, and $K_m$ is the total number of pixels per sample. For the MSE loss, we isolate the valid reconstruction targets using a binary reconstruction mask $\mathbf{V}_m \in \{0, 1\}^{(N_m, K_m)}$, over which the MSE below is normalized; the dense objectives that follow instead normalize over all $K_m$ pixels.

\noindent The masked squared error tensor is:
\begin{equation}
\mathbf{U}_{\text{MSE}, m} = \mathbf{V}_m \odot \left( \mathbf{P}_m - \mathbf{T}_m \right)^2
\end{equation}
where $\odot$ denotes element-wise multiplication. For a specific sample $i$ containing modality $m$, the reduced MSE scalar is obtained by averaging strictly over the masked elements:
\begin{equation}
L_{\text{MSE}, m}^{(i)} = \frac{1}{\sum_{j=1}^{K_m} \mathbf{V}_{m,j}^{(i)}} \sum_{j=1}^{K_m} \mathbf{U}_{\text{MSE}, m, j}^{(i)}
\end{equation}

\noindent In contrast, dense complementary objectives are computed over the complete spatial grid. Denoting a generic dense loss function as $f_{\text{dense}}$ (e.g., SSIM or FFT), $f_{\text{dense}}(\mathbf{P}^{(i)}, \mathbf{T}^{(i)})_j$ is its value at location $j$ (for SSIM, the local windowed comparison around $j$, not a pointwise one), and the scalar loss for sample $i$ is its global spatial mean:
\begin{equation}
L_{\text{dense}, m}^{(i)} = \frac{1}{K_m} \sum_{j=1}^{K_m} f_{\text{dense}}\left(\mathbf{P}^{(i)}, \mathbf{T}^{(i)}\right)_{m,j}
\end{equation}

The complete scalar loss $L_m^{(i)}$ for modality $m$ on sample $i$ is a weighted sum of these terms plus a modality weight $\alpha_m$, with $w_{\text{MSE}}, w_c$ fixed configuration weights and $C$ the set of additional dense objectives applied to modality $m$ (e.g., SSIM, FFT):
\begin{equation}
L_m^{(i)} = \alpha_m \left( w_{\text{MSE}} L_{\text{MSE}, m}^{(i)} + \sum_{c \in C} w_c L_{c, m}^{(i)} \right)
\end{equation}

\noindent Our three reduction strategies differ in how sample-level scalar losses $L_m^{(i)}$ are aggregated into the final batch loss $\mathcal{L}$. Let $N_{\text{val}}$ represent the number of samples in the batch with at least one available modality (i.e. $|M_i| > 0$), $M$ be the set of all active modalities present across the batch, and $M_i$ represent the subset of modalities available for a specific sample $i$.

\vspace{2pt} \noindent \textbf{Strategy A: No Balancing.}
The simplest reduction sums the per-modality losses within each sample before averaging over $N_{\text{val}}$:
\begin{equation}
\mathcal{L}_{\text{none}} = \frac{1}{N_{\text{val}}} \sum_{i=1}^{N_{\text{val}}} \left( \sum_{m \in M_i} L_{m}^{(i)} \right)
\end{equation}
This is the standard Empirical Risk Minimization formulation and treats every valid pixel of every valid sensor equally. In a sparse multi-sensor setting, samples with more co-occurring modalities naturally accumulate a higher total loss, and thus produce larger gradient magnitudes.

\vspace{2pt} \noindent \textbf{Strategy B: Modality Balancing.}
This strategy computes the mean loss per modality independently across its specific subset of $N_m$ valid samples, and sums the resulting averages across all active modalities in the batch:
\begin{equation}
\mathcal{L}_{\text{modality}} = \sum_{m \in M} \left( \frac{1}{N_m} \sum_{i=1}^{N_m} L_{m}^{(i)} \right)
\end{equation}
Each modality contributes exactly one unit of average error regardless of how many samples contain it, ensuring under-represented sensors receive consistent optimization focus. The risk is higher gradient variance for rare modalities, potentially destabilising training.

\vspace{2pt} \noindent \textbf{Strategy C: Sample Balancing.}
This formulation normalizes the intra-sample loss sum by its specific count of valid modalities $|M_i|$ before taking the average over the batch:
\begin{equation}
\mathcal{L}_{\text{sample}} = \frac{1}{N_{\text{val}}} \sum_{i=1}^{N_{\text{val}}} \left( \frac{1}{|M_i|} \sum_{m \in M_i} L_{m}^{(i)} \right)
\end{equation}
Every sample exerts equal gradient pull regardless of its modality count, but this penalizes information-rich samples, as a four-modality sample contributes the same magnitude as a single-modality one. \\

\subsection{Token Aggregation}
Existing architectures such as MultiMAE \cite{multimae} and subsequent works \cite{2023_4m, 2024_4m21} typically handle cross-modal interaction via independent decoding pathways, utilizing cross-attention modules to merge the global encoder context specifically in relation to the target modality being reconstructed. While this mechanism allows for flexible feature decoding, it defers the step of multi-modal alignment to the individual decoder heads rather than enforcing a unified representation earlier in the pipeline. Ideally, we would like an autonomous perception system to have physical sensor streams aligned at the encoder level, so that downstream features map to a more consistent embedding space regardless of which sparse subset of sensors provided the input. The aim of individual decoding heads should be a matter of simple modality-specific style reconstruction rather than also taking care of the fundamental alignment.

\vspace{2pt} \noindent We test the effect of an injected bottleneck between the encoder and the decoders as a substitute for the conventional cross-attention mechanism to enforce this behavior. Architecturally, this bottleneck is realized by introducing an MLP component that operates on the spatial grids produced by the fusion encoder.
The module directly concatenates the tokens of all active input streams by aligning them at the same spatial dimensions across different modalities. The tokens are then  projected back into a single embedding using a linear layer, producing a single shared spatial representation that preserves spatial correspondence. Unavailable tokens or missing modalities are replaced with a zero matrix, preventing unstructured tensor operations.

\section{Experimental Evaluation}
\label{sec:evaluation}

We evaluate on two benchmarks covering complementary sensor sets. SeeThroughFog (STF) \cite{bijelic2020seeing} is a multi-sensor AD dataset captured under four real weather conditions — clear, light fog, dense fog, and snow/rain — providing RGB, NIR, Stereo Depth, LiDAR, and radar streams. Model selection is always performed on the respective validation splits, never on the test sets, used for final reporting only; the full protocol is detailed in Appendix~\ref{sec:model_selection}. For mapping Radar and LiDAR to the 2D plane, we use the original code of \cite{bijelic2020seeing} and feed the mapped Depth channel to the fine-tuning training. For Radar, it assumes the 2D plane is orthogonal to the image plane, therefore invariant along the vertical axis. Stereo Depth is retrieved from the RGB stereo camera. STF is evaluated using the standard KITTI evaluation protocol, following \cite{bijelic2020seeing}. DSEC \cite{gehrig2021dsec, Gehrig3dv2021} is an Event-camera dataset for driving which includes Event camera recordings in harsh lighting conditions. The dataset is partially annotated with semantic segmentation masks~\cite{sun2022ess} (evaluated through mIoU) and fully annotated with object bounding boxes~\cite{gehrig2024low} (evaluated using the COCO evaluation metric). While DSEC is utilized during pretraining, the test set is withheld to maintain an unbiased evaluation. We report results for \textit{Our-ViT} and \textit{Our-MM}, which share the same in-domain pretraining procedure (hyperparameters in Appendix \ref{sec:appendix_pretraining}), differing only in their initialization: standard ViT weights and MultiMAE weights, respectively. We consider ViT \cite{dosovitskiy2020image}, MAE \cite{he2022masked}, DINO \cite{dino} and MultiMAE \cite{multimae} as baselines because they represent the most commonly used pretrained weights built upon the standard ViT architecture. Sharing the exact same backbone configuration allows for a direct evaluation of different pretraining methods.
Our evaluation results are summarized in Table~\ref{tab:stf_main} for STF, and in Table~\ref{tab:dsec_main} for DSEC. Evaluation results on additional datasets are provided in Appendix \ref{sec:extended_evaluation}.
\begin{table}[t]
\centering
\renewcommand{\arraystretch}{0.92}
\caption{%
  \textbf{Evaluation Results on STF.} We report 2D bounding box AP$_{R40}$ (\%) using the standard KITTI evaluation protocol.
  The reported scores represent the average across all object classes for the easy, moderate, and hard difficulty levels.
  C\,=\,Clear; LF\,=\,Light Fog; DF\,=\,Dense Fog; S\,=\,Snow/Rain.
  \textbf{Bold} indicates the best result. \underline{Underline} indicates the second best. This is a summarized version of Table \ref{tab:weather_single} in Appendix.
}
\label{tab:stf_main}
\scriptsize
\setlength{\tabcolsep}{0pt}
\providecommand{\dcw}{}
\newlength{\dcwlen}
\setlength{\dcwlen}{0.1025\columnwidth}
\begin{tabular}{@{}p{0.12\columnwidth}
  @{}>{\centering\arraybackslash}p{\dcwlen}
  @{}>{\centering\arraybackslash}p{\dcwlen}
  @{}>{\centering\arraybackslash}p{\dcwlen}
  @{}>{\centering\arraybackslash}p{\dcwlen}
  @{\ \ }|@{\ \ }
  >{\centering\arraybackslash}p{\dcwlen}
  @{}>{\centering\arraybackslash}p{\dcwlen}
  @{}>{\centering\arraybackslash}p{\dcwlen}
  @{}>{\centering\arraybackslash}p{\dcwlen}@{}}
\specialrule{.10em}{0pt}{0pt}
\specialrule{.04em}{2pt}{2pt}
& \textbf{C} & \textbf{LF} & \textbf{DF} & \textbf{S}
& \textbf{C} & \textbf{LF} & \textbf{DF} & \textbf{S} \\
\specialrule{.04em}{2pt}{0pt}
\noalign{\smallskip}
& \multicolumn{4}{c}{\textit{RGB}}
& \multicolumn{4}{c}{\textit{NIR}} \\
\noalign{\smallskip}
\specialrule{.04em}{0pt}{2pt}
ViT           & 61.91 & 63.58 & 54.98 & 60.43 & 64.98 & 65.09 & 57.21 & 63.00 \\
MAE           & 65.79 & 68.15 & 59.11 & 64.92 & 69.30 & 67.60 & 60.85 & 67.35 \\
DINO          & 65.25 & 70.93 & 60.27 & 65.45 & 68.83 & 68.94 & 61.59 & 66.43 \\
MultiMAE      & \underline{70.41} & \underline{70.98} & \underline{66.91} & \textbf{68.42} & \textbf{71.72} & \underline{71.59} & \underline{69.24} & \underline{68.61} \\
Our-ViT       & 69.43 & 68.94 & 61.15 & 65.54 & 70.81 & 70.52 & 65.12 & 68.42 \\
Our-MM        & \textbf{71.39} & \textbf{72.29} & \textbf{68.08} & \underline{67.97} & \underline{71.67} & \textbf{73.40} & \textbf{69.33} & \textbf{69.26} \\
\specialrule{.04em}{2pt}{0pt}
\noalign{\smallskip}
& \multicolumn{4}{c}{\textit{Depth}}
& \multicolumn{4}{c}{\textit{RGB+NIR}} \\
\noalign{\smallskip}
\specialrule{.04em}{0pt}{2pt}
ViT           & 37.08 & 29.69 & 11.22 & 31.65 & 70.16 & 71.07 & 61.44 & 69.10 \\
MAE           & 40.41 & 33.86 & 14.72 & 34.84 & 74.15 & 75.74 & 66.75 & 72.46 \\
DINO          & 38.73 & 33.48 &  9.93 & 34.07 & 73.04 & \underline{77.37} & 63.47 & 71.29 \\
MultiMAE      & 44.80 & \textbf{42.75} & 17.45 & 39.49 & \underline{76.03} & 76.47 & \underline{71.93} & \underline{74.61} \\
Our-ViT       & \underline{46.68} & \underline{41.48} & \underline{19.60} & \underline{40.98} & 75.08 & 76.77 & 69.71 & 73.92 \\
Our-MM        & \textbf{48.02} & 41.34 & \textbf{20.20} & \textbf{42.60} & \textbf{76.45} & \textbf{77.98} & \textbf{74.47} & \textbf{75.07} \\
\specialrule{.04em}{2pt}{0pt}
\noalign{\smallskip}
& \multicolumn{4}{c}{\textit{RGB+Depth}}
& \multicolumn{4}{c}{\textit{RGB+LiDAR}} \\
\noalign{\smallskip}
\specialrule{.04em}{0pt}{2pt}
ViT           & 63.77 & 65.00 & 49.33 & 62.31 & 64.87 & 64.81 & 56.30 & 61.46 \\
MAE           & 68.34 & 69.37 & 54.72 & 66.16 & 67.45 & 68.23 & 60.29 & 65.49 \\
DINO          & 67.26 & 70.18 & 53.09 & 65.25 & 66.43 & 67.67 & 54.68 & 64.05 \\
MultiMAE      & \underline{71.21} & \textbf{73.60} & \textbf{61.85} & \underline{69.65} & 69.81 & 70.33 & \textbf{66.85} & \underline{69.49} \\
Our-ViT       & 70.94 & 70.59 & 58.78 & 67.14 & \underline{70.02} & \underline{70.35} & 64.42 & 66.94 \\
Our-MM        & \textbf{72.46} & \underline{72.17} & \underline{61.07} & \textbf{69.95} & \textbf{72.52} & \textbf{73.25} & \underline{66.73} & \textbf{69.91} \\
\specialrule{.04em}{2pt}{0pt}
\noalign{\smallskip}
& \multicolumn{4}{c}{\textit{RGB+NIR+Depth}}
& \multicolumn{4}{c}{\textit{RGB+NIR+LiDAR}} \\
\noalign{\smallskip}
\specialrule{.04em}{0pt}{2pt}
ViT           & 70.94 & 72.71 & 58.26 & 70.67 & 71.18 & 72.45 & 62.55 & 70.12 \\
MAE           & 74.23 & 76.67 & 60.48 & 72.75 & 74.59 & 76.13 & 68.43 & 72.62 \\
DINO          & 72.83 & 76.41 & 60.25 & 71.74 & 72.95 & 77.04 & 63.06 & 71.35 \\
MultiMAE      & 74.86 & \underline{78.73} & \underline{67.53} & \underline{73.77} & \underline{75.51} & \textbf{79.41} & \underline{74.34} & \underline{75.18} \\
Our-ViT       & \underline{75.27} & \textbf{78.92} & 67.15 & 73.34 & 75.17 & 78.06 & 69.75 & 74.07 \\
Our-MM        & \textbf{77.13} & 78.51 & \textbf{68.83} & \textbf{75.57} & \textbf{76.80} & \underline{79.13} & \textbf{74.80} & \textbf{75.89} \\
\specialrule{.04em}{2pt}{2pt}
\specialrule{.10em}{0pt}{0pt}
\end{tabular}
\end{table}
\vspace{2pt} \noindent \textbf{Finding 1: MultiMAE is the strongest off-the-shelf baseline.}
Looking across nearly all configurations, a clear ranking emerges: \textit{ViT} weights perform worst, followed by \textit{MAE} and \textit{DINO} at a similar level. \textit{MultiMAE} outperforms almost all backbones, indicating that it is a very strong out-of-the-box baseline that transfers effectively to physical sensors it has never seen.

\vspace{2pt} \noindent \textbf{Finding 2: In-domain pretraining helps specifically when non-RGB modalities are included as inputs.}
\textit{Our-MM} manages to improve on MultiMAE frequently across all benchmarks, but the margins are narrow.
A recurring pattern across datasets is that the advantage of \textit{Our-MM} over \textit{MultiMAE} increases as soon as non-RGB modalities are included as inputs. For RGB-only inputs the margin is small, while as soon as we introduce Stereo Depth, LiDAR, LWIR, or Event data, in most settings \textit{Our-MM} improves incrementally on top of \textit{MultiMAE}. This suggests that the in-domain pretraining specifically improves the encoder's ability to integrate physically heterogeneous sensor streams.

\begin{table}[t]
\centering
\renewcommand{\arraystretch}{0.92}
\caption{%
\textbf{Evaluation Results on DSEC.} We report mean Intersection over Union (mIoU) for semantic segmentation and average precision thresholds ($mAP_{50:95}$, $mAP_{50}$) for object detection across isolated and fused modalities. \textbf{Bold} indicates the best result. \underline{Underline} indicates the second best.%
}
\label{tab:dsec_main}
\scriptsize
\setlength{\tabcolsep}{0pt}
\ifdefined\dcwlen\else\newlength{\dcwlen}\fi
\setlength{\dcwlen}{0.1358\columnwidth}
\begin{tabular}{@{}p{0.13\columnwidth}
  @{}>{\centering\arraybackslash}p{\dcwlen}
  @{\hspace{3pt}\textcolor{gray!60}{\vrule width 0.3pt}\hspace{3pt}}>{\centering\arraybackslash}p{\dcwlen}
  @{}>{\centering\arraybackslash}p{\dcwlen}
  @{\ \ }|@{\ \ }
  >{\centering\arraybackslash}p{\dcwlen}
  @{\hspace{3pt}\textcolor{gray!60}{\vrule width 0.3pt}\hspace{3pt}}>{\centering\arraybackslash}p{\dcwlen}
  @{}>{\centering\arraybackslash}p{\dcwlen}@{}}
\specialrule{.10em}{0pt}{0pt}
\specialrule{.04em}{2pt}{2pt}
& \multicolumn{1}{c}{\textit{Seg.}} & \multicolumn{2}{c}{\textit{Det.}}
& \multicolumn{1}{c}{\textit{Seg.}} & \multicolumn{2}{c}{\textit{Det.}} \\
\cmidrule(r){2-2}\cmidrule(lr){3-4}\cmidrule(r){5-5}\cmidrule(l){6-7}
& \textbf{mIoU} & $\textbf{mAP}_{\mathbf{50:95}}$ & $\textbf{mAP}_{\mathbf{50}}$
& \textbf{mIoU} & $\textbf{mAP}_{\mathbf{50:95}}$ & $\textbf{mAP}_{\mathbf{50}}$ \\
\specialrule{.04em}{2pt}{0pt}
\noalign{\smallskip}
& \multicolumn{3}{c}{\textit{RGB}}
& \multicolumn{3}{c}{\textit{Event}} \\
\noalign{\smallskip}
\specialrule{.04em}{0pt}{2pt}
ViT      & 54.45          & 35.50          & 63.48          & 42.79          & 28.20 & 49.86 \\
MAE      & 55.71          & 40.80          & 71.70          & 43.46          & 33.89 & 57.33 \\
DINO     & 54.29          & 37.37          & 66.66          & 43.49          & 32.30 & 55.26 \\
MultiMAE & \underline{62.61} & \underline{46.46} & \underline{77.11} & \underline{50.05} & \textbf{36.88} & \textbf{60.78} \\
Our-ViT  & 61.20          & 43.18          & 73.09          & 48.88          & 30.25 & 50.68 \\
Our-MM   & \textbf{64.25} & \textbf{46.77} & \textbf{77.24} & \textbf{52.22} & \underline{36.15} & \underline{58.91} \\
\specialrule{.04em}{2pt}{0pt}
\noalign{\smallskip}
& \multicolumn{3}{c}{\textit{RGB+Depth}}
& \multicolumn{3}{c}{\textit{RGB+Event}} \\
\noalign{\smallskip}
\specialrule{.04em}{0pt}{2pt}
ViT      & 46.83          & 36.32          & 65.58          & 32.48          & 39.47          & 62.83          \\
MAE      & 55.56          & 41.59          & 73.53          & 51.97          & 51.00          & 77.46          \\
DINO     & 40.86          & 38.38          & 68.45          & 40.02          & 44.24          & 69.24          \\
MultiMAE & 62.71 & \underline{46.60} & \underline{77.41} & \underline{63.34} & \underline{53.46} & \textbf{80.01} \\
Our-ViT  & \underline{62.89}          & 44.55          & 74.71          & 62.78          & 48.89          & 74.09          \\
Our-MM   & \textbf{64.73} & \textbf{48.05} & \textbf{79.60} & \textbf{65.84} & \textbf{54.47} & \underline{79.99} \\
\specialrule{.04em}{2pt}{0pt}
\noalign{\smallskip}
& \multicolumn{3}{c}{\textit{Event+Depth}}
& \multicolumn{3}{c}{\textit{RGB+Event+Depth}} \\
\noalign{\smallskip}
\specialrule{.04em}{0pt}{2pt}
ViT      & 43.00          & 37.28          & 61.37          & 33.74          & 44.58          & 70.64          \\
MAE      & 47.14          & \underline{41.01}          & \underline{65.83}          & 36.55          & 52.07          & 78.70          \\
DINO     & 42.45          & 39.07          & 62.97          & 41.94          & 47.37          & 72.72          \\
MultiMAE & 53.63 & 40.32 & 63.52 & \underline{63.92} & \underline{53.21} & \underline{79.23} \\
Our-ViT  & \underline{54.29}         & 38.28          & 61.62         & 63.13          & 51.02          & 77.33          \\
Our-MM   & \textbf{57.72} & \textbf{42.29} & \textbf{67.40} & \textbf{66.22} & \textbf{55.07} & \textbf{81.47} \\
\specialrule{.04em}{2pt}{2pt}
\specialrule{.10em}{0pt}{0pt}
\end{tabular}
\end{table}

On STF, performance degrades monotonically with harsher weather conditions. Dense fog is the condition that separates methods most clearly: while all models degrade, \textit{MultiMAE} and \textit{Our-MM} maintain a larger gap over \textit{ViT}, \textit{MAE}, and \textit{DINO}, suggesting that richer pretraining representations are more robust to severe visibility degradation. Furthermore, we observe the widest gaps when Stereo Depth and LiDAR modalities are present, suggesting that multimodal pretraining often helps with non-RGB sensors. NIR transfers surprisingly well from \textit{MultiMAE}'s pretraining, possibly because similar semantic structure generalizes across spectra. For RGB-only downstream tasks, in-domain pretraining adds marginal value.
Looking at DSEC, and in particular the Event modality, the gap between \textit{MultiMAE} and \textit{Our-MM} is also narrow, strongly indicating that generic multi-modal pretraining weights generalize to very different new modalities. Pretraining specifically on these modalities does incrementally improve overall quality by a few points.
Importantly, \textit{Our-ViT}, initialized from standard \textit{ViT} weights rather than \textit{MultiMAE} weights, already outperforms \textit{MAE} and \textit{DINO} across the board, showing that the in-domain pretraining data and objective contribute significantly regardless of initialization. The additional gain of \textit{Our-MM} over \textit{Our-ViT} appears attributable to the MultiMAE-style initialization, which provides a beneficial multimodal starting point for in-domain training. We further stress-test the \textit{RGB+Event+Depth} DSEC finetune to study its behaviour under sensor dropout and missing-modality conditions at inference time; the robustness to sensor dropout induced by the MAE strategy is confirmed, with \textit{Our-MM} degrading the least when sensors are unavailable (Appendix~\ref{sec:robustness}). We additionally compare performance across model scales (\textit{Base}, \textit{Small}, \textit{Tiny}) in Appendix~\ref{sec:model_scale}.

\vspace{2pt} \noindent \textbf{Finding 3: In-domain pretraining provides a faster convergence advantage.}
Our pretrained weights show an advantage in the early training steps, suggesting that in-domain initialization provides a better starting point for adapting to new sensor modalities. However, the gap narrows as training progresses. This hints that, for fine-tuning, the general ability to integrate information across modalities may matter more than the specificity of the pretrained representations. Once the network has been exposed to the target modality through fine-tuning, even \textit{MultiMAE}'s more generic representations prove more than sufficient. Examples of this behaviour can be observed in Figure \ref{fig:learn_curves}, Appendix \ref{sec:extended_evaluation}.

\subsection{Ablations}
\label{sec:ablations}
\begin{table}[t]
    \centering
    \caption{\textbf{Ablation studies} on different pretraining settings. We report mAP (mean $\pm$ std over 3 seeds) on downstream tasks for the STF and LLVIP datasets across various modality configurations. Bold indicates the selected configuration for the final pretrain.}
    \label{tab:ablations}
    \resizebox{\linewidth}{!}{
    \begin{tabular}{cp{4cm}cccc}
        \toprule
        & & \multicolumn{2}{c}{\textbf{STF}} & \multicolumn{2}{c}{\textbf{LLVIP}} \\
        \cmidrule(r){3-4} \cmidrule(l){5-6}
        \textbf{ID} & \textbf{Pretraining Setting} & \makecell{RGB + NIR + \\ Depth + Radar} & \makecell{NIR + Depth + \\ Radar} & \makecell{RGB + LWIR} & \makecell{LWIR} \\
        \midrule
        A0 & Baseline (\textit{Rand. init, MSE, Sample balancing, cross-attention, no semantic.}) & 35.77 \small{$\pm$0.12} & 30.78 \small{$\pm$0.34} & 62.03 \small{$\pm$0.43} & 61.81 \small{$\pm$0.20} \\
        \midrule
        \multicolumn{6}l{\textit{Initialization Weights}} \\
        A1 & ViT weights init. & 38.62 \small{$\pm$0.30} & 35.66 \small{$\pm$0.37} & 62.56 \small{$\pm$0.46} & 62.94 \small{$\pm$0.21} \\
        \textbf{A2} & \textbf{MultiMAE weights init.} & 39.01 \small{$\pm$0.18} & 36.67 \small{$\pm$0.62} & 63.68 \small{$\pm$0.46} & 62.99 \small{$\pm$0.41} \\
        \midrule
        \multicolumn{6}l{\textit{Reconstruction Losses}} \\
        \textbf{A3} & \textbf{MSE + SSIM Loss} & 37.62 \small{$\pm$0.30} & 34.98 \small{$\pm$0.14} & 63.04 \small{$\pm$0.81} & 62.19 \small{$\pm$0.45} \\
        A4 & MSE + FFT Loss & 35.35 \small{$\pm$0.31} & 32.71 \small{$\pm$0.12} & 61.12 \small{$\pm$0.22} & 61.56 \small{$\pm$0.68} \\
        \midrule
        \multicolumn{6}l{\textit{Loss Balancing}} \\
        A5 & No balancing & 37.49 \small{$\pm$0.34} & 34.41 \small{$\pm$0.23} & 62.45 \small{$\pm$0.31} & 62.33 \small{$\pm$0.47} \\
        \textbf{A6} & \textbf{Modality balancing} & 37.22 \small{$\pm$0.46} & 33.95 \small{$\pm$0.62} & 62.40 \small{$\pm$0.76} & 62.34 \small{$\pm$0.47} \\
        \midrule
        \textbf{A7} & \textbf{w/ Tokens Aggregation} & 37.69 \small{$\pm$0.19} & 34.78 \small{$\pm$0.55} & 63.16 \small{$\pm$0.15} & 62.49 \small{$\pm$0.39} \\
        \midrule
        \textbf{A8} & \textbf{w/ Semantic modality} & 38.50 \small{$\pm$0.27} & 35.40 \small{$\pm$0.19} & 64.39 \small{$\pm$0.35} & 63.17 \small{$\pm$0.95} \\
        \bottomrule
    \end{tabular}
    }
\end{table}

To determine the optimal pretraining setting, we examine the effect of different setups. Following the practice in \cite{multimae, 2023_4m}, we define a reference model and alternate one setting at a time for a reduced number of 400 epochs. We report the mean Average Precision (mAP) on selected downstream tasks across the STF and LLVIP datasets over 3 initializations with different seeds (Table \ref{tab:ablations}).

\vspace{2pt} \noindent \textbf{Reference Model.} The reference model encoder is a ViT-base-224-patch16 with random weight initialization (A0). It takes as input all real sensor-based modalities: RGB, LWIR, NIR, Event, and Stereo Depth. Modality-specific tokens are processed by modality-specific decoders with cross-attention. Finally, MSE Loss is computed for all modalities, sample balanced, and back-propagated with a constant learning rate of 2e-4.

\vspace{2pt} \noindent \textbf{Pretraining Weights.} To give the model a strong starting point and reduce computational costs, we test loading weights from existing models. We evaluate classic ViT weights (A1) and MultiMAE weights (A2). Initializing with MultiMAE weights yields the best performance overall, with the clearest gains when multiple modalities are present.

\vspace{2pt} \noindent \textbf{Loss Variations.} We examine the importance of additional losses that focus on reconstructing finer details, specifically SSIM loss (A3) and FFT Loss (A4). Adding SSIM loss improves results across almost all tasks, making it our choice for the final model. Conversely, FFT loss slightly degrades performance compared to the baseline.

\vspace{2pt} \noindent \textbf{Data Balancing.} We study the effects of the proposed balancing strategies: sample balancing (A0), no balancing (A5), and modality balancing (A6). At this reduced 400-epoch budget, the mean mAP differences between A5 and A6 are small and fall within one standard deviation across all configurations. We select A6 by design, which ensures equal representation across diverse sensors, preventing dominant modalities from overshadowing sparse ones.

\vspace{2pt} \noindent \textbf{Token Aggregation.} We evaluate the impact of using token aggregation (A7) to merge spatially overlapping features. We observe that incorporating this module leads to a performance gain over the baseline, though within noise on LWIR.

\vspace{2pt} \noindent \textbf{Semantics.} Similar to MultiMAE, we add semantics as a sixth modality (A8) to force the model to learn object-focused representations. This addition also improves results on all considered datasets. The final configuration is highlighted in bold in Table \ref{tab:ablations} and fully detailed in Appendix \ref{sec:appendix_pretraining}.

\section{Conclusion}
We asked which pretrained weights transfer best to physical sensor finetuning, and whether in-domain pretraining can outperform off-the-shelf solutions. The answer is conditional. \textit{MultiMAE} is the strongest out-of-the-box baseline in most settings, transferring effectively even to modalities unseen during pretraining. In-domain pretraining yields frequent improvements, but specifically when non-RGB sensors are involved: for RGB-only tasks, standard \textit{MultiMAE} weights remain difficult to beat.
Beyond final performance, in-domain weights converged faster in our runs, suggesting their primary value may lie in the cross-modal priors they provide rather than in low-level feature learning alone.

\vspace{8pt}
{\fontsize{8}{9}\selectfont\noindent\textbf{Acknowledgements.} This work has been supported by Innovation Fund Denmark under grant agreement No. 2105-00013A (SAVA) and the Novo Nordisk Foundation under grant agreement No. NNF25SA0104538 (Robotic Intercropping). HPC resources were provided by the Pioneer Centre for AI, DTU and the Novo Nordisk Foundation through access to the Gefion AI Supercomputer.

\noindent\textbf{Contributions.} \textit{Pretrain data curation:} Sandby, Montello. \textit{Finetune data curation:} Güldenring, Montello, Øhrstrøm. \textit{Algorithm design:} Montello, Güldenring, Sandby. \textit{Benchmark and ablations:} Güldenring, Montello. \textit{Writing:} Montello, Güldenring, Nalpantidis.  \textit{Supervision:} Nalpantidis.\par}

{
    \small
    \bibliographystyle{ieeenat_fullname}
    \bibliography{references}
}

\clearpage
\appendix

\twocolumn[
    \begin{center}
        \Large\bfseries Do Pretrained Backbones Transfer to Physical Sensors?\\A Systematic Study of Multimodal Initialization Across the Sensor Stack \\
        \vspace{0.5em}
        \large --- Supplementary Material --- \\
        \vspace{1.5em}
    \end{center}
]
\input{appendix/datasets}
\input{appendix/pretraining_hyp}
\input{appendix/finetuning_hyp}
\input{appendix/results}
\input{appendix/robustness}
\input{appendix/licensing}
\include{appendix/samples}
\end{document}

%% file: appendix/datasets.tex
\section{Datasets}
\label{sec:datasets}
This Appendix section provides a detailed overview of the 16 datasets selected to compile our multi-modal pretraining set: BDD100k \cite{Yu2018BDD100KAD}, Cityscapes \cite{Cordts2016Cityscapes}, DSEC \cite{gehrig2021dsec, Gehrig3dv2021}, ECMD \cite{10342726}, FMB \cite{liu2023segmif}, FR-T \cite{vertens2020heatnet}, KAIST-MS-Pedestrian \cite{7298706}, LLVIP \cite{jia2021llvip}, MFNet \cite{8206396}, MS2 \cite{shin2023deep}, NSAVP \cite{carmichael2025dataset}, NTU4DRadLM \cite{ZhangZhugeLiu2023ITSC}, RASMD \cite{JIN2026103872}, SMOD \cite{11146667}, STheReO \cite{9981857}, Vivid++ \cite{9760091}. Our collection favored datasets composed of dense 2D representations, excluding datasets with sparse modalities such as LiDAR or Radar. For this reason, datasets that heavily rely on these modalities (e.g. KITTI) have not been considered in the pretraining corpus, but are used for evaluation. We describe the specific preprocessing workflows implemented for each dataset and we outline the exact train, validation, and test split configurations used to maintain consistency across our evaluation pipeline.
\paragraph{BDD100k \cite{Yu2018BDD100KAD}} is an RGB-only large-scale driving dataset with a focus on highly diverse geographic, environmental, and weather conditions. The dataset also includes labels for a subset of the data on pixel-level segmentation, object detection, lane markings and other annotations. The dataset already provides train/validation/test splits. We preprocessed the images with vertical cropping, splitting in half the original horizontal resolution from $1280\times720$ to two non-overlapping images of $640\times720$.

\paragraph{Cityscapes \cite{Cordts2016Cityscapes}} is the de-facto standard benchmark when it comes to urban street scenes. It provides pixel-level semantic and instance segmentation for autonomous driving. The dataset is already composed of train/validation/test splits. We use the left camera for the RGB modality, while the pair of left and right RGB images is used to generate the Depth label through FoundationStereo \cite{wen_trepte_aribido_kautz_gallo_birchfield}, as described in Appendix \ref{sec:foundationstereo}. We cropped all RGB-Depth pairs keeping only the top \textit{775} pixels to remove the Mercedes logo present on the bottom portion of all the images. Furthermore, we cropped each image pair horizontally into partially overlapping patches of $640\times775$ resolution. We leave the images untouched in the finetuning for the semantic segmentation task.

\paragraph{DSEC \cite{gehrig2021dsec, Gehrig3dv2021}} is a stereo RGB-Event camera dataset for autonomous driving, focusing on high-dynamic-range and high-speed visual odometry and 3D perception. We synchronize and spatially align Event camera data with stereo RGB frames, and sample them at $5Hz$. Then camera rectification is applied, and we project the Event field-of-view onto the RGB images to ensure overlap. The Depth image is then generated through FoundationStereo. Finally, we partition the training portion to train and validation sets by migrating \textit{2\%} of the RGB-Event-Depth pairs.

\paragraph{ECMD \cite{10342726}} is an Event-centric multisensory driving dataset focusing on challenging scenarios. It contains RGB left-right pairs, Event and LWIR images. We generate frames from Event data, then we undistort the images, synchronize them with the RGB images, sampling at \textit{4Hz}. Subsequently, we generate the Depth with FoundationStereo and create the validation split by moving 2\% of the data. Since the test set is not provided, all images are split between training and validation. Finally we calculate the common Field of View among RGB-Event-LWIR-Depth pairs and crop/resize all the modalities to match.

\paragraph{FMB \cite{liu2023segmif}} is a dataset for semantic segmentation in the context of Thermal sensor fusion across varied time and weather conditions. It is composed of RGB-LWIR pairs already split between train/validation and test sets. Segmentation masks are also provided, and we use them for semantic segmentation finetuning.

\paragraph{Freiburg Thermal (FR-T) \cite{vertens2020heatnet}} is a multimodal RGB-LWIR dataset focusing on bridging the day-night gap in semantic segmentation. We started with cropping to a resolution of $1400\times650$ to remove border artifacts from the original images. We then sampled the set with 1Hz frequency and subsequently cropped the pairs into overlapping patches of $640\times650$.

\paragraph{KAIST-MS-Pedestrian \cite{7298706}} is a multispectral dataset containing aligned RGB and LWIR images, focusing on robust pedestrian detection across day and night conditions. We perform 6Hz sampling as the only preprocessing step.

\paragraph{LLVIP \cite{jia2021llvip}} is an RGB-LWIR dataset focusing on low-light vision for pedestrian detection in very dark, nighttime street environments. It is not strictly an AD dataset, but shows many of the features of urban scenarios. We kept the original train/val/test split and performed a horizontal overlapping crop for all pairs to reduce the resolution to $640\times1024$. The original image sizes are used for object detection finetuning.

\paragraph{MFNet \cite{8206396}} is an RGB-LWIR urban driving dataset for semantic segmentation of city scenes. The original dataset stores visible and Thermal information into a 4 channel single image. We decompose them and perform the splits according to the provided text files. The dataset is also used for testing the quality of multimodal downstream finetuning.

\paragraph{MS2 \cite{shin2023deep}} is a multi-spectral stereo dataset (RGB, NIR, LWIR, and LiDAR) focusing on adverse environmental conditions. The data is provided already rectified and synchronized. We downsample it to $5Hz$ and we allocate 2\% of the images into the validation set. No test set is provided. We perform spatial alignment for all sensors, projecting them to the RGB field of view, with subsequent crop to the common overlapping area. Depth labels are then generated through FoundationStereo. Finally, the images are cropped into overlapping patches to reduce dimensionality, with a final resolution of $640\times384$.

\paragraph{NSAVP \cite{carmichael2025dataset}} is a dataset featuring stereo Event and stereo LWIR cameras for place recognition, localization, and mapping. We started from the provided HDF5 files. The RGB and LWIR modalities are then synced and downsampled to $5Hz$. Unmatched files are deleted. 2\% of the images are used to create the validation split. We then proceed to generate the Events following the method in Section~\ref{subsec:modalities_datasets}, by lazily slicing 50ms windows around specific image timestamps and accumulating them into multi-bin voxel grids. Images are undistorted and cropped into a common field of view, sets of images with no Events are filtered out and Depth maps are generated from left-right RGB pairs using FoundationStereo.

\paragraph{NTU4DRadLM \cite{ZhangZhugeLiu2023ITSC}} is a 4D radar-centric multi-modal dataset which also includes RGB and LWIR modalities, focusing on localization and mapping (SLAM) tasks. We start by extracting, undistorting and synchronizing (50ms window) RGB and LWIR images from the provided ROS bag files. The data is downsampled to $5Hz$ and 2\% of the total data is kept aside for validation. The RGB-LWIR pairs are then projected and cropped into a common field of view.

\paragraph{RASMD \cite{JIN2026103872}} is a dataset which features synchronized RGB and Short-Wave Infrared (SWIR) images. The target task is object detection in adverse weather conditions (fog, rain). The processing in this case prepares the RASMD dataset by sampling down to $4Hz$ matched pairs of RGB and SWIR images, shuffling them, and distributing the data into 98\% training and 2\% validation splits. Note that SWIR is aggregated in the Near-Infrared (NIR) bucket in our pipeline, given their high similarity in wavelength.

\paragraph{SMOD \cite{11146667}} is an RGB-LWIR dataset focusing on object detection with different degrees of occlusion. In this case the dataset has simply been split between train and validation, with a 2\% ratio.

\paragraph{STheReO \cite{9981857}} is a stereo RGB-LWIR dataset with the addition of navigation sensors, for the tasks of Thermal odometry, mapping, and SLAM in visually degraded environments. We process the dataset by debayering, rectifying and cropping the stereo RGB images as well as LWIR. All pairs get projected to a common field of view and we proceed to synchronize right and LWIR pairs within a 50ms window. We also reduce the frequency from the original $10Hz$ to $3.33Hz$. Finally we calculate the Depth from the left/right image pairs using FoundationStereo.

\paragraph{Vivid++ \cite{9760091}} is a comprehensive dataset (including Event, LWIR, RGB and LiDAR data) created to support robust visual SLAM and place recognition under illumination changes. We make use only of the driving scenes. We extract samples from ROS bags, synchronize them among sensors and downsample the files from $30Hz$ to $5Hz$. Events are generated according to the procedure in Section~\ref{subsec:modalities_datasets}. The Depth map is generated from left-right image pairs using FoundationStereo. 2\% of the dataset is kept for validation, the rest is used for training.

\section{Model Selection Protocol}
\label{sec:model_selection}
To avoid any risk of test-set leakage, we use a strict separation between validation and test splits throughout both pretraining and finetuning. During pretraining, checkpoint selection is based exclusively on the aggregated validation split described in Table~\ref{tab:dataset_overview} (i.e., the held-out 2\% partitions listed per dataset, or the dataset's own provided validation split when available); we monitor the total reconstruction loss on this split and select the checkpoint with the lowest validation loss for all downstream finetuning. During finetuning, the same principle applies per downstream benchmark: for STF, we select the best checkpoint based on validation-set AP$_{R40}$ following the dataset's official validation partition; for DSEC, model selection is based on validation mIoU (segmentation) and validation mAP (detection) using DSEC's provided validation split. Test splits — STF's held-out test partition and the withheld DSEC test set described in Section~\ref{sec:evaluation} — are used strictly for final reported evaluation numbers in Tables~\ref{tab:stf_main} and~\ref{tab:dsec_main}, and are never used for hyperparameter tuning, early stopping, or checkpoint selection at any stage.

\section{Depth and Semantic maps generation}
\label{sec:foundationstereo}
We make use of FoundationStereo \cite{wen_trepte_aribido_kautz_gallo_birchfield} for the generation of the Depth image starting from a stereo RGB pair. FoundationStereo is a large-scale foundation model for stereo Depth estimation designed to achieve robust zero-shot generalization across diverse domains without requiring per-domain fine-tuning. For efficiency, all predictions are generated from the ViT-small checkpoint. We performed the inference at half-resolution with respect to the original image size (i.e. for Cityscapes with resolution $2048\times775$, the model works with resolution $1024\times388$). Subsequently we rescale the image to its original size through linear interpolation. We used the default hyperparameter \textit{20} for \textit{valid\_iters} when generating the ONNX. The resulting output is stored with float16 precision.\\

We employ Mask2Former \cite{cheng2022masked} to generate pixel-level semantic segmentation labels for all the RGB images (independently from already available masks, for consistency throughout the dataset). Mask2Former is a universal image segmentation framework that unifies semantic, instance, and panoptic segmentation under a single mask classification architecture. We utilize the \textit{large Swin Transformer} backbone variant pretrained on the \textit{Cityscapes} dataset to maximize zero-shot label quality across our urban driving scenes. The resulting discrete semantic labels are stored in uint8.

%% file: appendix/pretraining_hyp.tex
\section{Pretraining Details}
\label{sec:appendix_pretraining}
This section outlines the hyperparameter settings, architecture modifications, and token allocation strategies used during the pretraining phase. Our solution was pretrained on 8$\times$NVIDIA DGX H100 GPUs (including ablations and the final weights).

\begin{table}[ht]
\centering
\caption{Pretraining hyperparameters. These settings apply to the primary Base model variant.}
\label{tab:pretrain_hyperparams}
\resizebox{\columnwidth}{!}{%
\begin{tabular}{ll}
\toprule
\textbf{Configuration} & \textbf{Value} \\
\midrule
\multicolumn{2}{c}{\textit{Optimization}} \\
\midrule
\textit{Optimizer} & AdamW \\
\textit{Base learning rate} & 2e-4 \\
\textit{Weight decay} & 0.05 \\
\textit{Optimizer momentum} & $\beta_1=0.9, \beta_2=0.95$ \\
\textit{Learning rate schedule} & Cosine decay \\
\textit{Minimum learning rate} & 0 \\
\textit{Warmup epochs} & 40 \\
\textit{Warmup starting LR} & 1e-6 \\
\textit{Epochs} & 1600 \\
\textit{Effective batch size} & 512 \\
\textit{Gradient accumulation steps} & 1 \\
\textit{Precision} & BF16-mixed \\
\midrule
\multicolumn{2}{c}{\textit{Architecture \& Tokens}} \\
\midrule
\textit{Input image resolution} & $224 \times 224$ \\
\textit{Patch size} & $16 \times 16$ \\
\textit{Input tokens per sample} & 98 \\
\textit{Loss functions} & MSE + SSIM (Images), CE (Segm) \\
\textit{Loss balancing} & Modality balancing\\
\midrule
\multicolumn{2}{c}{\textit{Data Augmentation}} \\
\midrule
\textit{Geometric (Shared)} & RandomResizedCrop (0.2, 1.0) \\
& Horizontal Flip ($p=0.5$) \\
\textit{Photometric (Thermal)} & MinMax + CLAHE (2.0, [8,8]) \\
\bottomrule
\end{tabular}%
}
\end{table}

\subsection{Hyperparameters}
Our pretraining configuration builds on the setup used in MultiMAE, adapted for a highly sparse, heterogeneous multi-modal dataset. We train two models (with different initializations) for 1600 epochs with an effective batch size of 512. The network is optimized using AdamW with a base learning rate of 2e-4, scaled through a linear warmup and cosine decay schedule. Table \ref{tab:pretrain_hyperparams} provides the complete list of training parameters.

\subsection{Loss Formulation and Balancing}
The total reconstruction loss is a balanced sum across the available modalities for each sample. For image-like modalities (RGB, Thermal, Depth, and accumulated Event frames), we apply a combined loss of Mean Squared Error (MSE) and Structural Similarity Index (SSIM) with equal weight. The MSE is calculated exclusively over the masked (unvisited) patches to direct gradients toward cross-modal reconstruction. The SSIM is computed over the full spatial grid. For the semantic segmentation modality, we compute a standard cross-entropy loss over the 19 Cityscapes classes.

\subsection{Masking and Token Budget}
Token allocation is determined dynamically per sample. Given a fixed input budget of 98 tokens, we sample a Dirichlet distribution with a uniform concentration parameter ($\alpha=1$) to partition the budget among the modalities present in the current sample, while ensuring that every available modality receives at least one input token. The number of output tokens (reconstruction targets) is symmetrically set to 98. This continuous random allocation ensures the fusion encoder sees different modality proportions at every step.

%% file: appendix/finetuning_hyp.tex
\section{Appendix: Fine-Tuning Details}
\label{sec:appendix_finetuning}

\begin{table*}[t]
\centering
\caption{Fine-tuning hyperparameters for semantic segmentation and object detection tasks.}
\label{tab:combined_hyperparams}
\begin{tabular*}{\textwidth}{@{\extracolsep{\fill}}p{3.5cm}p{6.3cm}p{6.3cm}}
\toprule
\textbf{Configuration} & \textbf{Semantic Segmentation} & \textbf{Object Detection} \\
\midrule
\multicolumn{3}{c}{\textit{Optimization}} \\
\midrule
\textit{Optimizer} & AdamW & AdamW \\
\textit{Base learning rate} & 1e-3\newline 2e-4 (DSEC, FMB MM, MFNet) & 1e-4 \\
\textit{Backbone learning rate} & 2e-4\newline 1e-4 (DSEC, FMB MM, MFNet) & 1e-5 \\
\textit{Weight decay} & 0.05 & 0.1 \\
\textit{Optimizer momentum} & $\beta_1=0.9, \beta_2=0.999$ & $\beta_1=0.9, \beta_2=0.999$ \\
\textit{Learning rate schedule} & Cosine decay & None \\
\textit{Minimum learning rate} & 1e-6 & — \\
\textit{Warmup epochs} & 5 (BDD100K)\newline 10 (Cityscapes, FMB RGB)\newline 20 (DSEC)\newline 25 (FMB MM, MFNet) & 5 \\
\textit{Gradient clip value} & 5.0\newline 1.0 (DSEC, FMB MM, MFNet) & 0.1 \\
\textit{Epochs} & 100 (BDD100K)\newline 150 (Cityscapes)\newline 200 (DSEC)\newline 250 (FMB, MFNet) & 50\newline 30 (DSEC)\newline 80 (STF) \\
\textit{Batch size} & 64 & 64 \\
\midrule
\multicolumn{3}{c}{\textit{Architecture \& Regularization}} \\
\midrule
\textit{Head type} & Segmenter & Cascade R-CNN \\
\textit{Layer decay} & 0.75 & 1.0 \\
\textit{Drop path rate} & 0.1 & 0.1 \\
\midrule
\multicolumn{3}{c}{\textit{Data Augmentation}} \\
\midrule
\textit{Photometric} & Color Jitter 0.4 & Color Jitter 0.4 \\
\textit{Geometric} & LSJ [0.75, 2.0]\newline LSJ [0.8, 1.5] (DSEC, FMB, MFNet) & LSJ[0.5, 2.0] \\
\bottomrule
\end{tabular*}
\end{table*}
This section outlines the implementation details used to fine-tune the evaluated ViT backbone weights. All experiments share a common optimization framework but are adapted in the learning schedules and resolutions to the specific characteristics and scale of each dataset and task.

\subsection{Multimodal Adapter Architecture}
Each input modality receives an independent 2D convolutional patch projection layer with a kernel and stride of 16. We assign independent spatial positional embeddings and a learnable classification token for each sensor type. During the forward pass, positional embeddings are bicubically interpolated to match the current input resolution. When initializing standard RGB-only pretrained models (i.e. ViT, MAE, DINO and some instances of MultiMAE) for non-RGB inputs, we average the pretrained RGB patch projection weights across the channel dimension and expand them to the required channel count. Classification tokens are concatenated along the sequence dimension and processed jointly by the shared transformer encoder. \\
The fusion part of the model happens in between the encoder and the decoder: tokens are concatenated along the channel dimension and merged via a single linear fusion layer. Subsequently they are passed to the respective task head.

\subsection{Input Adapter Initialization}
\label{sec:adapter_init}
For ViT, MAE, and DINO, all non-RGB modalities (Depth, LWIR, NIR, Event) are initialized by averaging the pretrained RGB patch-projection kernel across the channel dimension and tiling it to the target channel count, while positional embeddings are cloned from the pretrained RGB embedding and the classification token is shared across all modalities; notably, MultiMAE follows the identical averaging procedure for every non-RGB modality, except Depth where the natively pretrained adapter is used. For \textit{Our-ViT} and \textit{Our-MM} we load modality-specific adapters, positional embeddings, and modality embeddings directly from our in-domain pretrained checkpoint.

\subsection{Task Heads}
\textbf{Semantic Segmentation.} We use a lightweight Segmenter decoder \cite{strudel2021segmenter} consisting of learnable class prototype embeddings and a 2-layer transformer encoder (pre-norm, GELU, 8 heads). The final segmentation masks are generated via a dot product between the refined patch and class representations, followed by bilinear upsampling.
\\
\textbf{Object Detection.} We use a Cascade R-CNN head~\cite{cai2018cascadercnn}, which is trained with increasing IoU thresholds leading to a higher emphasis on precise bounding box prediction.

\subsection{Dataset-Specific Hyperparameters}
We optimize all models using AdamW with mixed precision (BF16) and a weight decay of 0.05 for segmentation and 0.1 for detection. We apply a layer-wise learning rate decay of 0.75 across the backbone blocks, except for the STF benchmark where it is set to 1.0.

Table~\ref{tab:combined_hyperparams} details the exact learning rates, input resolutions, and training schedules applied to each dataset.

%% file: appendix/results.tex
\begin{table*}[b]
\centering
\small
\caption{Quantitative evaluation across single-modal and multimodal downstream tasks. Performance is measured via mean Intersection over Union (mIoU) for semantic segmentation on Cityscapes, BDD100K, FMB, and MFNet, and via Average Precision ($mAP_{50:95}$ and $mAP_{50}$) for object detection on LLVIP and KITTI. Best results are in \textbf{bold}, and second best are \underline{underlined}.}
\label{tab:wacv_results}
\begin{tabular}{llll|cccccc}
\toprule
\textbf{Dataset} & \textbf{Task} & \textbf{Modality} & \textbf{Metric} & \textbf{ViT} & \textbf{MAE} & \textbf{Dino} & \textbf{MultiMAE} & \textbf{Our-ViT} & \textbf{Our-MM} \\
\midrule
\textbf{Cityscapes} & Sem. Seg. & RGB & mIoU & 70.26 & 65.74 & 69.45 & \textbf{73.88} & 66.74 & \underline{70.75} \\
\midrule
\textbf{BDD100K} & Sem. Seg. & RGB & mIoU & 53.01 & 50.31 & 52.33 & \textbf{54.95} & 52.41 & \underline{53.34} \\
\midrule
\textbf{LLVIP} & Obj. Det. & RGB & $\text{mAP}_{50:95}$ & 46.15 & \underline{48.11} & 45.34 & 47.97 & 45.53 & \textbf{49.04} \\
 & & & $\text{mAP}_{50}$ & 86.01 & \textbf{86.91} & 83.97 & 86.79 & 84.00 & \underline{86.83} \\
\cmidrule{3-10}
 & & LWIR & $\text{mAP}_{50:95}$ & 54.89 & \textbf{60.49} & 57.10 & \underline{60.33} & 57.01 & 59.52 \\
 & & & $\text{mAP}_{50}$ & 91.98 & \textbf{95.74} & 93.76 & \underline{94.67} & 92.00 & 93.91 \\
\cmidrule{3-10}
 & & RGB - LWIR & $\text{mAP}_{50:95}$ & 57.04 & \textbf{60.81} & 58.35 & \underline{60.71} & 58.55 & 59.27 \\
 & & & $\text{mAP}_{50}$ & 94.05 & \underline{95.11} & 94.86 & \textbf{95.38} & 92.31 & 93.23 \\
\midrule
\textbf{KITTI} & Obj. Det. & RGB & $\text{mAP}_{50:95}$ & 30.61 & 36.66 & 33.72 & \underline{37.94} & 37.81 & \textbf{38.85} \\
 & & & $\text{mAP}_{50}$ & 54.68 & 62.48 & 58.88 & \underline{63.64} & 63.35 & \textbf{64.88} \\
\cmidrule{3-10}
 & & Stereo Depth & $\text{mAP}_{50:95}$ & \underline{28.24} & 26.10 & \textbf{29.93} & 27.97 & 28.03 & 27.78 \\
 & & & $\text{mAP}_{50}$ & \underline{52.59} & 48.84 & \textbf{55.21} & 52.21 & 51.13 & 51.29 \\
\cmidrule{3-10}
 & & RGB - Stereo Depth & $\text{mAP}_{50:95}$ & 34.06 & 36.92 & 35.21 & \underline{38.21} & 38.12 & \textbf{39.26} \\
 & & & $\text{mAP}_{50}$ & 59.94 & 63.32 & 60.78 & 63.89 & \underline{64.22} & \textbf{66.22} \\
\midrule
\textbf{FMB} & Sem. Seg. & RGB & mIoU & 45.18 & 44.93 & 46.13 & \textbf{53.59} & 51.06 & \underline{53.10} \\
\cmidrule{3-10}
 & & RGB - LWIR & mIoU & 39.44 & 44.94 & 44.21 & 50.51 & \underline{53.91} & \textbf{54.68} \\
\midrule
\textbf{MFNet} & Sem. Seg. & RGB & mIoU & 46.11 & 56.39 & 49.86 & \textbf{66.18} & 60.07 & \underline{63.41} \\
\cmidrule{3-10}
 & & RGB - LWIR & mIoU & 54.04 & 58.18 & 49.08 & 67.77 & \underline{64.78} & \textbf{69.33} \\
\bottomrule
\end{tabular}
\end{table*}

\providecommand{\ms}[2]{#1$_{\pm#2}$}
\begin{table*}[!t]
\centering
\caption{%
  \textbf{Multi-seed Evaluation Results on STF.} 2D bounding box AP$_{R40}$ (\%) using the standard KITTI evaluation protocol,
  averaged across all object classes for the easy, moderate, and hard difficulty levels.
  We report mean\,$\pm$\,std over seeds 42, 43 and 44.
  C\,=\,Clear; LF\,=\,Light Fog; DF\,=\,Dense Fog; S\,=\,Snow/Rain.
  \textbf{Bold} indicates the best result. \underline{Underline} indicates the second best.
}
\label{tab:stf_main_multiseed}
\footnotesize
\setlength{\tabcolsep}{5pt}
\begin{tabular*}{\textwidth}{@{\extracolsep{\fill}}l cccc c cccc@{}}
\specialrule{.10em}{0pt}{0pt}
\specialrule{.04em}{2pt}{2pt}
& \textbf{C} & \textbf{LF} & \textbf{DF} & \textbf{S} &
& \textbf{C} & \textbf{LF} & \textbf{DF} & \textbf{S} \\
\specialrule{.04em}{2pt}{0pt}
\noalign{\smallskip}
& \multicolumn{4}{c}{\textit{RGB}} & & \multicolumn{4}{c}{\textit{NIR}} \\
\noalign{\smallskip}
\cmidrule(lr){2-5}\cmidrule(lr){7-10}
MultiMAE & \ms{\underline{69.99}}{0.56} & \ms{\underline{70.98}}{0.51} & \ms{\underline{67.08}}{0.72} & \ms{\textbf{68.46}}{0.06}
         & & \ms{\underline{71.86}}{0.21} & \ms{\underline{71.19}}{0.61} & \ms{\underline{68.49}}{1.23} & \ms{\underline{68.78}}{0.27} \\
Our-ViT  & \ms{69.20}{0.73} & \ms{69.48}{0.99} & \ms{62.91}{1.75} & \ms{65.44}{1.12}
         & & \ms{70.53}{0.28} & \ms{70.47}{0.20} & \ms{65.53}{1.33} & \ms{68.46}{0.19} \\
Our-MM   & \ms{\textbf{70.94}}{0.59} & \ms{\textbf{71.62}}{0.96} & \ms{\textbf{68.32}}{1.24} & \ms{\underline{67.98}}{0.08}
         & & \ms{\textbf{72.22}}{0.51} & \ms{\textbf{73.43}}{0.16} & \ms{\textbf{69.09}}{0.32} & \ms{\textbf{69.47}}{0.36} \\
\specialrule{.04em}{2pt}{0pt}
\noalign{\smallskip}
& \multicolumn{4}{c}{\textit{Depth}} & & \multicolumn{4}{c}{\textit{RGB+NIR}} \\
\noalign{\smallskip}
\cmidrule(lr){2-5}\cmidrule(lr){7-10}
MultiMAE & \ms{44.57}{0.32} & \ms{42.09}{0.69} & \ms{18.13}{0.60} & \ms{39.25}{0.29}
         & & \ms{\underline{75.67}}{0.69} & \ms{77.56}{0.98} & \ms{\underline{72.30}}{1.11} & \ms{\underline{75.01}}{0.44} \\
Our-ViT  & \ms{\underline{47.18}}{0.63} & \ms{\underline{42.68}}{1.04} & \ms{\underline{18.93}}{0.64} & \ms{\underline{41.13}}{0.13}
         & & \ms{75.36}{0.32} & \ms{\underline{77.62}}{0.97} & \ms{70.12}{0.78} & \ms{74.07}{0.15} \\
Our-MM   & \ms{\textbf{47.89}}{0.34} & \ms{\textbf{43.35}}{1.78} & \ms{\textbf{19.75}}{0.45} & \ms{\textbf{42.08}}{0.47}
         & & \ms{\textbf{76.52}}{0.22} & \ms{\textbf{79.13}}{1.00} & \ms{\textbf{73.87}}{0.52} & \ms{\textbf{75.14}}{0.56} \\
\specialrule{.04em}{2pt}{0pt}
\noalign{\smallskip}
& \multicolumn{4}{c}{\textit{RGB+Depth}} & & \multicolumn{4}{c}{\textit{RGB+LiDAR}} \\
\noalign{\smallskip}
\cmidrule(lr){2-5}\cmidrule(lr){7-10}
MultiMAE & \ms{\underline{71.01}}{0.19} & \ms{\underline{72.95}}{0.92} & \ms{\textbf{62.03}}{0.54} & \ms{\underline{69.26}}{0.35}
         & & \ms{69.99}{0.48} & \ms{\underline{71.18}}{0.79} & \ms{\textbf{66.84}}{0.46} & \ms{\underline{69.40}}{0.53} \\
Our-ViT  & \ms{70.75}{0.21} & \ms{71.15}{1.00} & \ms{59.16}{0.47} & \ms{67.12}{0.26}
         & & \ms{\underline{70.01}}{0.06} & \ms{69.57}{0.96} & \ms{63.40}{1.38} & \ms{67.13}{0.42} \\
Our-MM   & \ms{\textbf{72.44}}{0.14} & \ms{\textbf{73.35}}{1.11} & \ms{\underline{61.78}}{0.79} & \ms{\textbf{70.12}}{0.24}
         & & \ms{\textbf{71.98}}{0.47} & \ms{\textbf{72.72}}{0.56} & \ms{\underline{64.99}}{1.51} & \ms{\textbf{69.71}}{0.27} \\
\specialrule{.04em}{2pt}{0pt}
\noalign{\smallskip}
& \multicolumn{4}{c}{\textit{RGB+NIR+Depth}} & & \multicolumn{4}{c}{\textit{RGB+NIR+LiDAR}} \\
\noalign{\smallskip}
\cmidrule(lr){2-5}\cmidrule(lr){7-10}
MultiMAE & \ms{75.09}{0.20} & \ms{77.61}{1.05} & \ms{66.47}{0.94} & \ms{\underline{74.09}}{0.29}
         & & \ms{\underline{75.64}}{0.35} & \ms{77.94}{1.37} & \ms{\underline{72.57}}{1.68} & \ms{\underline{74.99}}{0.59} \\
Our-ViT  & \ms{\underline{75.42}}{0.16} & \ms{\textbf{79.16}}{1.13} & \ms{\underline{67.33}}{0.61} & \ms{74.00}{0.57}
         & & \ms{75.33}{0.17} & \ms{\underline{78.21}}{0.62} & \ms{70.36}{0.91} & \ms{74.45}{0.38} \\
Our-MM   & \ms{\textbf{77.03}}{0.15} & \ms{\textbf{79.16}}{0.56} & \ms{\textbf{68.86}}{1.74} & \ms{\textbf{75.75}}{0.58}
         & & \ms{\textbf{77.09}}{0.34} & \ms{\textbf{79.69}}{0.66} & \ms{\textbf{73.89}}{0.98} & \ms{\textbf{75.98}}{0.18} \\
\specialrule{.04em}{2pt}{2pt}
\specialrule{.10em}{0pt}{0pt}
\end{tabular*}
\end{table*}
\section{Extended Evaluation}
\label{sec:extended_evaluation}
Here we report the full set of experiments conducted, including the full list of results for the SeeThroughFog (STF) \cite{bijelic2020seeing} dataset, as well as the evaluation on standard single and multimodal datasets: Cityscapes\cite{Cordts2016Cityscapes}, BDD100K\cite{Yu2018BDD100KAD}, LLVIP \cite{jia2021llvip}, KITTI \cite{Geiger2012CVPR}, FMB \cite{liu2023segmif} and MFNet \cite{8206396}.
Counting first and second-place finishes across all datasets, modalities, weather, and difficulty combinations, \textit{Our-MM} achieves the highest number of top-ranked results, with \textit{MultiMAE} as the most frequent runner-up. This aggregate view confirms that the observed ranking is consistent and not an artifact of any particular evaluation condition.

For the STF dataset, we present further observations, considering the full scale of Table \ref{tab:weather_single}. \textit{Our-MM} degrades most slowly across nearly all modalities. On Stereo Depth, for instance, the drop from clear-easy to dense-fog-easy is substantially smaller for \textit{Our-MM} (55.56 → 22.00, $\sim$60\%) than for \textit{MultiMAE} (51.87 → 19.17, $\sim$63\%) and far smaller than for \textit{ViT} (43.79 → 12.58, $\sim$71\%). This relative robustness under distribution shift is a practically important property for real-world AD systems.

When using Radar-only, the AP is near zero across all models — the best single value across the entire table is 0.76 (\textit{Our-MM}, dense fog easy). We believe Radar point clouds projected to 2D image grids are too sparse to provide meaningful patch-level features for a ViT backbone.\\
Concerning weather conditions, unlike fog, which degrades performance monotonically from light to dense, snow/rain conditions typically sit between light and dense fog rather than below them in terms of results. This is physically plausible: snow and rain introduce different kinds of sensor noise than fog — primarily occlusion and reflection artifacts rather than uniform visibility loss — and their effect on Depth and Thermal sensors differs from their effect on RGB. This non-monotonic behavior is consistent across all models and modalities.\\
We observe also that \textit{Our-MM} degrades less than \textit{MultiMAE} on Stereo Depth and LiDAR under fog, while the gap between them is smaller on RGB and NIR. This is consistent with the fact that our pretraining corpus includes the Depth modality generated from physically acquired Stereo Depth, whereas \textit{MultiMAE}'s Depth is derived from a monocular model.

Table \ref{tab:wacv_results} shows the backbone comparison across other single and multimodal datasets. Considering standard RGB-only benchmarks, \textit{MultiMAE} leads clearly — 73.88 mIoU on Cityscapes and 54.95 on BDD100K — with \textit{Our-MM} second but noticeably below. These are the configurations most similar to \textit{MultiMAE}'s original pretraining domain: dense, photometric, ImageNet-adjacent data. In-domain pretraining on physical sensor data does not close this gap.\\
When we consider FMB and MFNet, \textit{MultiMAE} also wins or ties on RGB-only segmentation, but \textit{Our-MM} takes the lead on RGB+LWIR — 54.68 vs. 50.51 on FMB, and 69.33 vs. 67.77 on MFNet. The MultiMAE-style initialization combined with in-domain physical sensor pretraining specifically improves multi-sensor fusion, while single-modal photometric tasks still favor the original \textit{MultiMAE} weights.\\
Curiously, on LLVIP object detection, \textit{MAE} outperforms all networks on both RGB and LWIR single-modal inputs, while \textit{DINO} and \textit{ViT} seem to perform very well on the Stereo Depth only modality of KITTI.

\section{Multi-seed Evaluation}
\label{sec:app_multiseed}

To evaluate the sensitivity of our results to random initialization, we repeated the fine-tuning on STF for the three strongest models in Table~\ref{tab:stf_main}: MultiMAE, Our-ViT and Our-MM.
Each configuration is trained with seeds 42, 43 and 44 under the same protocol as the main experiments.
Table~\ref{tab:stf_main_multiseed} reports the mean and standard deviation of AP$_{R40}$ for each modality combination and weather condition.

The multi-seed results confirm the trends of the single-seed evaluation.
Our-MM outperforms MultiMAE in 29 of the 32 settings; MultiMAE remains ahead only in RGB under snow/rain and in RGB+Depth and RGB+LiDAR under dense fog.
The variance across seeds is low, with standard deviations mostly below 1\,AP.
Dense fog is the least stable condition on average, with deviations of up to 1.75\,AP.
In this condition, and for the RGB and RGB+NIR inputs, part of the margin over MultiMAE falls within seed variance.
The largest and most consistent gains are obtained in clear and snow/rain conditions with inputs that include Depth or LiDAR.
Among the unimodal settings, Our-ViT outperforms MultiMAE only with Depth input, indicating that the multimodal pretraining of Our-MM accounts for most of the improvement.
\begin{table*}[t]
\centering
\caption{Average Precision (AP) at 40 recall points (R40) averaged across all object classes, evaluated under four weather conditions: clear, light fog, dense fog, and snow/rain. Each condition reports three difficulty levels --- E (Easy), M (Moderate), and H (Hard) --- defined by the KITTI benchmark based on object size, occlusion level, and truncation degree, with Hard being the most challenging. Results are reported for single-modality and multi-modality input configurations across six backbone/pretraining methods: ViT (vanilla Vision Transformer), MAE (Masked Autoencoder), DINO (self-distillation with no labels), MultiMAE (multi-modal MAE baseline), and our proposed models Our-ViT and Our-MM. Best results per condition and difficulty are in \textbf{bold}, second best are \underline{underlined}.}
\label{tab:weather_single}
\resizebox{\linewidth}{!}{
\begin{tabular}{p{3cm}lcccccccccccc}
\toprule
\multirow{2}{*}{\textbf{Modality}} & \multirow{2}{*}{\textbf{Model}} & \multicolumn{3}{c}{\textbf{Clear}} & \multicolumn{3}{c}{\textbf{Light Fog}} & \multicolumn{3}{c}{\textbf{Dense Fog}} & \multicolumn{3}{c}{\textbf{Snow/Rain}} \\
\cmidrule(lr){3-5} \cmidrule(lr){6-8} \cmidrule(lr){9-11} \cmidrule(lr){12-14}
 & & E & M & H & E & M & H & E & M & H & E & M & H \\
\midrule
\multirow{6}{=}{\textbf{RGB}} & ViT & 68.50 & 60.55 & 56.69 & 68.92 & 62.33 & 59.49 & 59.38 & 54.03 & 51.52 & 67.29 & 58.72 & 55.29 \\
 & MAE & 72.05 & 64.47 & 60.84 & 72.79 & 67.23 & 64.42 & 63.32 & 58.11 & 55.90 & 71.35 & 63.44 & 59.98 \\
 & DINO & 71.38 & 64.12 & 60.25 & \underline{75.48} & 70.10 & 67.21 & 64.68 & 59.37 & 56.77 & 72.22 & 63.92 & 60.20 \\
 & MultiMAE & \underline{76.00} & \underline{69.39} & \underline{65.85} & 74.59 & \underline{70.43} & \underline{67.91} & \underline{70.47} & \underline{66.22} & \underline{64.05} & \textbf{74.10} & \textbf{66.99} & \textbf{64.16} \\
 & Our-ViT & 75.02 & 68.30 & 64.96 & 72.71 & 68.32 & 65.80 & 64.77 & 60.15 & 58.54 & 71.00 & 64.37 & 61.26 \\
 & Our-MM & \textbf{77.12} & \textbf{70.18} & \textbf{66.86} & \textbf{76.30} & \textbf{71.55} & \textbf{69.03} & \textbf{71.50} & \textbf{67.43} & \textbf{65.30} & \underline{73.47} & \underline{66.84} & \underline{63.60} \\
\midrule
\multirow{6}{=}{\textbf{NIR}} & ViT & 72.32 & 63.46 & 59.17 & 70.02 & 64.38 & 60.86 & 60.88 & 56.94 & 53.81 & 70.09 & 61.50 & 57.41 \\
 & MAE & 76.60 & 67.64 & 63.66 & 72.28 & 66.99 & 63.52 & 64.23 & 60.66 & 57.65 & 74.02 & 65.89 & 62.13 \\
 & DINO & 76.46 & 67.27 & 62.75 & 73.38 & 68.37 & 65.06 & 64.21 & 61.63 & 58.92 & 73.30 & 65.01 & 60.99 \\
 & MultiMAE & \textbf{78.55} & \textbf{70.30} & \underline{66.32} & \underline{75.74} & \underline{71.15} & \underline{67.87} & \underline{71.31} & \textbf{69.32} & \textbf{67.09} & \underline{74.73} & \underline{67.50} & \underline{63.61} \\
 & Our-ViT & 77.44 & 69.24 & 65.75 & 74.60 & 70.07 & 66.90 & 68.26 & 64.86 & 62.24 & 74.62 & 67.17 & 63.48 \\
 & Our-MM & \underline{77.97} & \underline{70.24} & \textbf{66.79} & \textbf{77.41} & \textbf{73.17} & \textbf{69.63} & \textbf{71.78} & \underline{69.31} & \underline{66.91} & \textbf{75.21} & \textbf{68.13} & \textbf{64.44} \\
\midrule
\multirow{6}{=}{\textbf{Stereo Depth}} & ViT & 43.79 & 35.28 & 32.18 & 33.49 & 28.67 & 26.90 & 12.58 & 10.95 & 10.14 & 37.02 & 30.13 & 27.79 \\
 & MAE & 47.48 & 38.54 & 35.20 & 38.12 & 32.65 & 30.82 & 16.43 & 14.34 & 13.39 & 40.83 & 33.13 & 30.55 \\
 & DINO & 45.79 & 36.69 & 33.71 & 37.50 & 32.39 & 30.56 & 10.81 & 9.73 & 9.24 & 39.85 & 32.37 & 30.00 \\
 & MultiMAE & 51.87 & 42.80 & 39.73 & \textbf{47.32} & \textbf{41.54} & \textbf{39.40} & 19.17 & 17.04 & 16.13 & 45.69 & 37.62 & 35.16 \\
 & Our-ViT & \underline{53.97} & \underline{44.50} & \underline{41.58} & \underline{45.85} & \underline{40.28} & \underline{38.30} & \underline{21.53} & \underline{19.18} & \underline{18.10} & \underline{47.16} & \underline{39.17} & \underline{36.62} \\
 & Our-MM & \textbf{55.56} & \textbf{45.82} & \textbf{42.68} & 45.74 & 40.14 & 38.13 & \textbf{22.00} & \textbf{19.79} & \textbf{18.80} & \textbf{49.14} & \textbf{40.74} & \textbf{37.93} \\
\midrule
\multirow{6}{=}{\textbf{Radar}} & ViT & 0.43 & 0.33 & 0.29 & 0.15 & 0.13 & 0.14 & 0.24 & 0.19 & 0.17 & 0.18 & 0.16 & 0.15 \\
 & MAE & 0.28 & 0.24 & 0.22 & 0.21 & 0.15 & 0.14 & 0.25 & 0.20 & 0.20 & 0.17 & 0.15 & 0.14 \\
 & DINO & 0.50 & 0.38 & 0.34 & 0.25 & 0.19 & \underline{0.18} & 0.41 & 0.32 & 0.30 & 0.28 & 0.23 & 0.22 \\
 & MultiMAE & \textbf{0.71} & \textbf{0.52} & \textbf{0.49} & 0.34 & 0.27 & \textbf{0.27} & 0.41 & 0.36 & 0.33 & \underline{0.34} & \textbf{0.28} & \textbf{0.27} \\
 & Our-ViT & 0.60 & 0.45 & \underline{0.40} & \underline{0.40} & \underline{0.29} & \textbf{0.27} & \underline{0.57} & \underline{0.44} & \textbf{0.43} & \textbf{0.37} & \underline{0.27} & \underline{0.25} \\
 & Our-MM & \underline{0.63} & \underline{0.47} & 0.38 & \textbf{0.41} & \textbf{0.30} & \textbf{0.27} & \textbf{0.76} & \textbf{0.49} & \underline{0.41} & \underline{0.34} & 0.26 & 0.24 \\
\midrule
\multirow{6}{=}{\textbf{LiDAR}} & ViT & 39.73 & 31.12 & 28.21 & 26.95 & 22.10 & 20.30 & 12.99 & 11.04 & 10.28 & 31.58 & 25.10 & 22.70 \\
 & MAE & 40.49 & 32.06 & 28.95 & 28.87 & 23.75 & 21.67 & 13.29 & 11.51 & 10.63 & 32.17 & 25.37 & 23.03 \\
 & DINO & 40.83 & 32.19 & 29.08 & 31.17 & 26.04 & 24.07 & \underline{14.75} & 12.57 & 11.68 & 33.83 & 26.81 & 24.43 \\
 & MultiMAE & 46.33 & 37.51 & 34.38 & 35.60 & 30.41 & 28.27 & 14.74 & 12.96 & 12.18 & 38.10 & 30.92 & 28.55 \\
 & Our-ViT & \underline{46.65} & \underline{38.07} & \underline{34.98} & \underline{36.92} & \underline{31.96} & \underline{30.00} & 14.66 & \underline{13.03} & \underline{12.30} & \textbf{39.85} & \textbf{32.70} & \textbf{30.20} \\
 & Our-MM & \textbf{47.12} & \textbf{38.29} & \textbf{35.11} & \textbf{38.50} & \textbf{33.31} & \textbf{31.28} & \textbf{16.49} & \textbf{14.54} & \textbf{13.67} & \underline{39.36} & \underline{32.08} & \underline{29.61} \\
\midrule
\multirow{6}{=}{\textbf{RGB - NIR}} & ViT & 76.71 & 68.93 & 64.83 & 75.06 & 70.76 & 67.40 & 65.30 & 61.01 & 58.01 & 75.25 & 67.98 & 64.07 \\
 & MAE & 80.12 & 73.11 & 69.21 & 79.47 & 75.29 & 72.45 & 70.91 & 66.42 & 62.92 & 78.39 & 71.48 & 67.50 \\
 & DINO & 79.61 & 72.00 & 67.51 & \underline{81.37} & \underline{77.02} & \underline{73.72} & 67.49 & 62.83 & 60.10 & 77.65 & 69.98 & 66.24 \\
 & MultiMAE & \underline{81.77} & \underline{75.02} & \underline{71.30} & 80.48 & 76.08 & 72.86 & \underline{75.38} & \underline{71.40} & \underline{69.01} & \underline{80.08} & \underline{73.41} & \underline{70.35} \\
 & Our-ViT & 80.52 & 74.10 & 70.61 & 80.63 & 76.39 & 73.30 & 72.77 & 69.34 & 67.01 & 79.35 & 73.07 & 69.35 \\
 & Our-MM & \textbf{81.85} & \textbf{75.50} & \textbf{72.01} & \textbf{81.53} & \textbf{77.74} & \textbf{74.66} & \textbf{77.42} & \textbf{74.18} & \textbf{71.82} & \textbf{80.37} & \textbf{73.99} & \textbf{70.86} \\
\midrule
\multirow{6}{=}{\textbf{RGB - Stereo Depth}} & ViT & 70.74 & 62.25 & 58.31 & 70.16 & 63.98 & 60.86 & 53.08 & 48.78 & 46.14 & 69.32 & 60.56 & 57.04 \\
 & MAE & 74.58 & 67.09 & 63.36 & 73.94 & 68.52 & 65.64 & 58.11 & 54.45 & 51.61 & 72.54 & 64.57 & 61.36 \\
 & DINO & 73.88 & 65.83 & 62.07 & 74.73 & 69.41 & 66.41 & 56.97 & 52.45 & 49.84 & 71.79 & 63.81 & 60.16 \\
 & MultiMAE & \underline{77.19} & \underline{69.70} & \underline{66.73} & \textbf{77.83} & \textbf{72.81} & \textbf{70.16} & \textbf{65.25} & \textbf{61.28} & \textbf{59.02} & \underline{75.42} & \underline{68.37} & \underline{65.16} \\
 & Our-ViT & 76.80 & 69.57 & 66.45 & 74.40 & 70.11 & 67.27 & 62.45 & 58.03 & 55.85 & 72.51 & 65.91 & 63.01 \\
 & Our-MM & \textbf{78.07} & \textbf{71.28} & \textbf{68.03} & \underline{76.01} & \underline{71.84} & \underline{68.66} & \underline{63.98} & \underline{60.50} & \underline{58.74} & \textbf{75.51} & \textbf{68.60} & \textbf{65.73} \\
\midrule
\multirow{6}{=}{\textbf{RGB - Radar}} & ViT & 69.11 & 61.18 & 57.15 & 69.01 & 62.89 & 59.65 & 59.84 & 54.56 & 51.77 & 68.65 & 59.76 & 56.10 \\
 & MAE & 73.77 & 66.37 & 62.55 & 73.34 & 68.29 & 65.20 & 66.43 & 61.27 & 58.26 & 71.35 & 63.75 & 60.17 \\
 & DINO & 71.12 & 63.37 & 59.33 & 73.65 & 68.03 & 64.66 & 61.17 & 55.39 & 52.72 & 69.85 & 61.69 & 57.78 \\
 & MultiMAE & 75.35 & \underline{69.04} & \underline{65.86} & 72.20 & 67.30 & 64.72 & \textbf{71.37} & \textbf{67.34} & \textbf{65.42} & \textbf{74.99} & \textbf{68.25} & \textbf{64.80} \\
 & Our-ViT & \underline{75.74} & 68.99 & 65.66 & \textbf{74.64} & \textbf{69.89} & \textbf{67.03} & 70.19 & 66.19 & 63.82 & 71.67 & 65.10 & 61.88 \\
 & Our-MM & \textbf{76.35} & \textbf{69.60} & \textbf{66.56} & \underline{74.47} & \underline{69.83} & \underline{66.98} & \underline{71.09} & \underline{66.49} & \underline{64.31} & \underline{73.46} & \underline{66.94} & \underline{63.58} \\
\bottomrule
\end{tabular}
}
\end{table*}
\begin{table*}[t]
\centering
\caption*{(Continued)}
\label{tab:weather_dual}
\resizebox{\linewidth}{!}{
\begin{tabular}{p{3cm}lcccccccccccc}
\toprule
\multirow{2}{*}{\textbf{Modality}} & \multirow{2}{*}{\textbf{Model}} & \multicolumn{3}{c}{\textbf{Clear}} & \multicolumn{3}{c}{\textbf{Light Fog}} & \multicolumn{3}{c}{\textbf{Dense Fog}} & \multicolumn{3}{c}{\textbf{Snow/Rain}} \\
\cmidrule(lr){3-5} \cmidrule(lr){6-8} \cmidrule(lr){9-11} \cmidrule(lr){12-14}
 & & E & M & H & E & M & H & E & M & H & E & M & H \\
\midrule
\multirow{6}{=}{\textbf{RGB - LiDAR}} & ViT & 71.60 & 63.54 & 59.47 & 69.33 & 63.90 & 61.21 & 60.44 & 55.61 & 52.85 & 68.19 & 59.91 & 56.27 \\
 & MAE & 73.41 & 66.31 & 62.63 & 72.66 & 67.50 & 64.53 & 64.26 & 59.51 & 57.09 & 72.01 & 63.94 & 60.52 \\
 & DINO & 72.76 & 65.24 & 61.28 & 72.17 & 66.81 & 64.03 & 58.96 & 53.80 & 51.28 & 70.56 & 62.60 & 59.00 \\
 & MultiMAE & 75.50 & 68.59 & 65.35 & 74.24 & \underline{69.86} & \underline{66.89} & \textbf{70.45} & \textbf{66.16} & \underline{63.94} & \underline{75.01} & \underline{68.34} & \underline{65.11} \\
 & Our-ViT & \underline{75.71} & \underline{68.89} & \underline{65.45} & \underline{74.43} & 69.84 & 66.79 & 67.96 & 63.86 & 61.44 & 72.40 & 65.99 & 62.44 \\
 & Our-MM & \textbf{78.16} & \textbf{71.35} & \textbf{68.04} & \textbf{76.90} & \textbf{72.75} & \textbf{70.10} & \underline{69.93} & \underline{66.09} & \textbf{64.16} & \textbf{75.55} & \textbf{68.64} & \textbf{65.54} \\
\midrule
\multirow{6}{=}{\textbf{NIR - LiDAR}} & ViT & 74.70 & 65.57 & 61.22 & 71.35 & 66.21 & 62.83 & 60.12 & 56.40 & 53.66 & 70.63 & 62.20 & 58.30 \\
 & MAE & 77.37 & 69.23 & 65.36 & 72.69 & 68.40 & 65.63 & 60.77 & 58.47 & 55.43 & 73.79 & 66.61 & 62.81 \\
 & DINO & 76.94 & 67.77 & 63.43 & 73.25 & 68.50 & 65.03 & 60.86 & 59.03 & 55.94 & 72.99 & 64.78 & 60.95 \\
 & MultiMAE & \underline{78.81} & \underline{70.75} & \underline{67.02} & \underline{75.98} & \underline{71.15} & \underline{68.27} & \underline{68.61} & \textbf{66.38} & \textbf{64.14} & \underline{75.73} & \underline{69.09} & \underline{65.56} \\
 & Our-ViT & 78.21 & 69.99 & 66.28 & 75.50 & 70.88 & 67.98 & \textbf{69.10} & \underline{65.88} & \underline{63.41} & 75.30 & 67.44 & 64.06 \\
 & Our-MM & \textbf{79.67} & \textbf{71.88} & \textbf{68.05} & \textbf{76.61} & \textbf{71.83} & \textbf{68.58} & 68.01 & 64.23 & 61.87 & \textbf{76.63} & \textbf{69.30} & \textbf{65.73} \\
\midrule
\multirow{6}{=}{\textbf{RGB - NIR - Stereo Depth}} & ViT & 77.78 & 69.46 & 65.59 & 76.80 & 72.34 & 68.98 & 62.60 & 57.29 & 54.90 & 76.82 & 69.30 & 65.88 \\
 & MAE & 80.51 & 73.09 & 69.10 & 80.85 & 76.20 & 72.96 & 64.14 & 60.07 & 57.22 & 78.98 & 71.58 & 67.68 \\
 & DINO & 79.18 & 71.53 & 67.78 & 80.74 & 75.92 & 72.57 & 63.23 & 60.33 & 57.18 & 77.95 & 70.54 & 66.72 \\
 & MultiMAE & 80.24 & 73.87 & 70.48 & \underline{82.58} & \textbf{78.44} & 75.17 & \underline{70.81} & \underline{67.20} & \underline{64.58} & \underline{79.16} & \underline{72.98} & \underline{69.16} \\
 & Our-ViT & \underline{80.94} & \underline{74.02} & \underline{70.85} & \textbf{82.88} & \underline{78.41} & \textbf{75.48} & 70.03 & 66.90 & 64.53 & 78.44 & 72.59 & 68.99 \\
 & Our-MM & \textbf{82.33} & \textbf{75.96} & \textbf{73.11} & 82.53 & 77.69 & \underline{75.32} & \textbf{71.17} & \textbf{68.92} & \textbf{66.40} & \textbf{80.60} & \textbf{74.42} & \textbf{71.70} \\
\midrule
\multirow{6}{=}{\textbf{RGB - NIR - Radar}} & ViT & 78.08 & 70.23 & 66.13 & 78.72 & 73.42 & 69.95 & 67.87 & 62.84 & 59.97 & 76.73 & 68.58 & 64.92 \\
 & MAE & 80.10 & 72.81 & 68.86 & 80.24 & 75.70 & 72.36 & 71.74 & 67.51 & 64.77 & 77.96 & 70.97 & 67.44 \\
 & DINO & 79.10 & 71.24 & 67.06 & 80.49 & 76.28 & 73.33 & 69.22 & 64.78 & 62.07 & 76.93 & 69.39 & 65.50 \\
 & MultiMAE & \underline{81.29} & \underline{74.47} & \underline{71.14} & 81.86 & 77.61 & 74.62 & \textbf{78.42} & \textbf{74.02} & \textbf{71.95} & \underline{80.26} & \underline{74.21} & \underline{70.78} \\
 & Our-ViT & 81.28 & 74.27 & 70.81 & \underline{83.00} & \underline{78.36} & \underline{75.65} & 72.68 & 69.26 & 66.79 & 79.34 & 73.54 & 70.05 \\
 & Our-MM & \textbf{82.36} & \textbf{75.95} & \textbf{72.83} & \textbf{83.25} & \textbf{79.05} & \textbf{76.34} & \underline{77.07} & \underline{73.35} & \underline{70.87} & \textbf{80.79} & \textbf{75.22} & \textbf{71.88} \\
\midrule
\multirow{6}{=}{\textbf{RGB - NIR - Stereo Depth - Radar}} & ViT & 77.92 & 70.07 & 66.10 & 77.70 & 72.44 & 69.23 & 62.09 & 57.43 & 54.39 & 77.21 & 69.25 & 65.70 \\
 & MAE & 80.44 & 72.65 & 69.10 & 78.56 & 74.30 & 71.28 & 63.46 & 58.91 & 55.96 & 78.51 & 71.06 & 67.38 \\
 & DINO & 79.76 & 72.25 & 68.64 & 81.31 & 76.47 & 73.37 & 67.63 & 62.67 & 59.85 & 78.36 & 70.92 & 67.09 \\
 & MultiMAE & 80.46 & 73.75 & 70.69 & 82.01 & 77.40 & 74.27 & 69.77 & 65.43 & 63.25 & 79.77 & 73.57 & \underline{70.37} \\
 & Our-ViT & \underline{81.80} & \underline{75.02} & \underline{71.49} & \underline{82.60} & \underline{78.30} & \underline{75.19} & \underline{70.74} & \textbf{67.81} & \textbf{65.75} & \underline{80.00} & \underline{73.81} & 70.06 \\
 & Our-MM & \textbf{82.23} & \textbf{75.63} & \textbf{71.98} & \textbf{84.35} & \textbf{80.22} & \textbf{77.42} & \textbf{71.55} & \underline{67.66} & \underline{65.08} & \textbf{81.21} & \textbf{74.65} & \textbf{71.48} \\
\midrule
\multirow{6}{=}{\textbf{RGB - NIR - LiDAR}} & ViT & 77.66 & 69.97 & 65.91 & 76.71 & 71.85 & 68.78 & 66.87 & 61.77 & 59.02 & 76.34 & 68.73 & 65.29 \\
 & MAE & 80.66 & 73.36 & 69.76 & 80.26 & 75.48 & 72.65 & 72.36 & 67.79 & 65.15 & 78.74 & 71.32 & 67.80 \\
 & DINO & 79.52 & 71.70 & 67.62 & 81.16 & 76.61 & 73.36 & 67.33 & 62.23 & 59.62 & 77.64 & 70.02 & 66.38 \\
 & MultiMAE & \underline{81.19} & \underline{74.32} & \underline{71.01} & \textbf{83.47} & \textbf{78.91} & \underline{75.85} & \underline{77.59} & \underline{73.88} & \underline{71.56} & \underline{80.07} & \underline{74.30} & \underline{71.18} \\
 & Our-ViT & 80.98 & 74.03 & 70.51 & 81.89 & 77.42 & 74.87 & 72.92 & 69.44 & 66.89 & 79.57 & 73.19 & 69.46 \\
 & Our-MM & \textbf{82.11} & \textbf{75.72} & \textbf{72.58} & \underline{82.82} & \underline{78.69} & \textbf{75.88} & \textbf{77.71} & \textbf{74.47} & \textbf{72.21} & \textbf{80.76} & \textbf{75.21} & \textbf{71.69} \\
\midrule
\multirow{6}{=}{\textbf{RGB - NIR - LiDAR - Radar}} & ViT & 78.75 & 70.43 & 66.75 & 79.56 & 74.37 & 70.90 & 69.73 & 64.58 & 61.86 & 77.17 & 69.71 & 65.79 \\
 & MAE & 80.81 & 73.72 & 69.67 & 79.60 & 74.49 & 71.64 & 68.81 & 64.28 & 61.62 & 78.33 & 71.33 & 67.76 \\
 & DINO & 79.23 & 71.32 & 67.34 & 80.05 & 75.66 & 72.19 & 67.11 & 63.51 & 60.27 & 78.29 & 71.13 & 67.50 \\
 & MultiMAE & \underline{81.55} & \underline{74.81} & \underline{71.98} & \underline{82.02} & 77.86 & 75.06 & \underline{76.47} & \underline{72.58} & \underline{70.47} & \textbf{81.41} & \textbf{74.91} & \textbf{71.55} \\
 & Our-ViT & 80.43 & 73.99 & 70.47 & 82.00 & \underline{77.91} & \underline{75.19} & 71.08 & 67.91 & 65.52 & 79.60 & 73.33 & 69.72 \\
 & Our-MM & \textbf{81.88} & \textbf{75.43} & \textbf{72.43} & \textbf{83.23} & \textbf{78.91} & \textbf{75.89} & \textbf{77.91} & \textbf{73.66} & \textbf{71.85} & \underline{80.92} & \underline{74.65} & \underline{71.33} \\
\midrule
\multirow{6}{=}{\textbf{NIR - Stereo Depth - Radar}} & ViT & 74.17 & 65.18 & 61.20 & 69.76 & 64.07 & 61.17 & 49.92 & 45.99 & 43.56 & 72.05 & 63.25 & 59.54 \\
 & MAE & 77.51 & 68.81 & 64.92 & 72.51 & 67.51 & 64.66 & 56.02 & 52.57 & 49.73 & 73.71 & 65.81 & 62.00 \\
 & DINO & 75.47 & 67.09 & 62.90 & 74.37 & 69.20 & 65.95 & 54.87 & 52.89 & 50.45 & 72.58 & 63.99 & 60.38 \\
 & MultiMAE & \textbf{79.32} & \underline{70.44} & \underline{67.10} & 76.61 & 72.81 & 69.39 & 62.45 & 59.79 & 57.33 & 75.97 & 68.22 & 64.87 \\
 & Our-ViT & 78.38 & 70.16 & 66.84 & \underline{78.38} & \underline{73.65} & \underline{70.09} & \underline{66.06} & \underline{63.95} & \underline{61.44} & \underline{76.44} & \underline{69.21} & \underline{65.51} \\
 & Our-MM & \underline{78.95} & \textbf{71.34} & \textbf{67.74} & \textbf{81.13} & \textbf{76.57} & \textbf{73.06} & \textbf{67.33} & \textbf{64.91} & \textbf{62.57} & \textbf{76.95} & \textbf{69.33} & \textbf{65.78} \\
\bottomrule
\end{tabular}
}
\end{table*}

%% file: appendix/robustness.tex
\clearpage

\section{Sensor Robustness and Missing Modalities}
\label{sec:robustness}
\begin{table*}[t]
\centering
\renewcommand{\arraystretch}{0.92}
\caption{%
  \textbf{Sensor Dropout Robustness on DSEC.} We report mIoU (\%) for semantic segmentation and $\text{mAP}_{50:95}$ / $\text{mAP}_{50}$ (\%) for object detection when the model receives only the listed subset of modalities at inference time, simulating sensor dropout or failure. \textbf{Bold} indicates the best result per column; \underline{underline} indicates the second best.
}
\label{tab:robustness_combined}
\scriptsize
\resizebox{\linewidth}{!}{
\begin{tabular}{lcccccccccccc}
\cmidrule[\heavyrulewidth]{1-10}
& \multicolumn{3}{c}{\textbf{RGB}} & \multicolumn{3}{c}{\textbf{Depth}} & \multicolumn{3}{c}{\textbf{Event}} & & & \\
\cmidrule(lr){2-4} \cmidrule(lr){5-7} \cmidrule(lr){8-10}
\textbf{Model} & mIoU & $\text{AP}_{50:95}$ & $\text{AP}_{50}$ & mIoU & $\text{AP}_{50:95}$ & $\text{AP}_{50}$ & mIoU & $\text{AP}_{50:95}$ & $\text{AP}_{50}$ & & & \\
\cmidrule{1-10}
ViT      & 18.40 & 17.59 & 35.70 & 1.42 & 0.42 & 0.50 & 4.99 & 0.80 & 1.26 & & & \\
MAE      & 33.67 & 21.05 & 40.92 & \textbf{9.72} & 0.00 & 0.00 & 5.67 & 1.15 & 1.83 & & & \\
DINO     & 27.98 & 17.29 & 31.41 & \underline{9.22} & 0.42 & 0.50 & 3.02 & 1.05 & 1.65 & & & \\
MultiMAE & 54.23 & 29.90 & \underline{58.74} & 2.04 & 2.53 & 4.21 & \textbf{21.33} & \underline{2.28} & 3.82 & & & \\
Our-ViT  & \underline{55.92} & \textbf{34.63} & \textbf{64.63} & 2.58 & \underline{9.81} & \underline{15.52} & \underline{11.17} & 2.21 & \textbf{4.13} & & & \\
Our-MM   & \textbf{59.20} & \underline{30.39} & 55.17 & 5.73 & \textbf{11.14} & \textbf{18.87} & 5.28 & \textbf{2.41} & \underline{4.10} & & & \\
\midrule
& \multicolumn{3}{c}{\textbf{Depth+Event}} & \multicolumn{3}{c}{\textbf{RGB+Depth}} & \multicolumn{3}{c}{\textbf{RGB+Event}} & \multicolumn{3}{c}{\textbf{All}} \\
\cmidrule(lr){2-4} \cmidrule(lr){5-7} \cmidrule(lr){8-10} \cmidrule(lr){11-13}
\textbf{Model} & mIoU & $\text{AP}_{50:95}$ & $\text{AP}_{50}$ & mIoU & $\text{AP}_{50:95}$ & $\text{AP}_{50}$ & mIoU & $\text{AP}_{50:95}$ & $\text{AP}_{50}$ & mIoU & $\text{AP}_{50:95}$ & $\text{AP}_{50}$ \\
\midrule
ViT      & 12.25 & 11.87 & 18.38 & 34.50 & 32.26 & 60.75 & 18.12 & 23.73 & 41.59 & 33.75 & 44.58 & 70.64 \\
MAE      & 14.43 & 7.27 & 11.95 & 34.96 & 38.88 & 71.17 & 34.08 & 27.78 & 45.17 & 36.55 & 52.07 & 78.70 \\
DINO     & 14.29 & 5.36 & 8.73 & 41.27 & 34.88 & 63.02 & 29.06 & 23.59 & 37.34 & 41.94 & 47.37 & 72.72 \\
MultiMAE & 29.93 & \textbf{20.87} & \textbf{34.14} & 60.18 & \underline{40.92} & \underline{72.68} & \underline{59.28} & 39.47 & 66.90 & \underline{63.92} & \underline{53.21} & \underline{79.23} \\
Our-ViT  & \textbf{38.01} & 17.61 & 27.78 & \underline{60.37} & 40.85 & 71.90 & 59.19 & \textbf{44.86} & \textbf{71.36} & 63.13 & 51.02 & 77.33 \\
Our-MM   & \underline{37.97} & \underline{20.07} & \underline{32.72} & \textbf{63.28} & \textbf{42.96} & \textbf{75.42} & \textbf{63.54} & \underline{44.72} & \underline{68.88} & \textbf{66.22} & \textbf{55.07} & \textbf{81.47} \\
\bottomrule
\end{tabular}
}
\end{table*}

A model that only performs well when every sensor is present is of limited use on a real vehicle, where a lens can fog up, a connector can fail, or an Event camera can simply saturate under strong sunlight. To probe this, we take the DSEC-finetuned checkpoint \textit{RGB+Event+Depth} from Section~\ref{sec:evaluation} and re-run inference while progressively dropping modalities, replacing the dropped modalities with black images. Table~\ref{tab:robustness_combined} reports detection and segmentation performance under these conditions. The isolated-sensor columns are the most telling: with Depth or Event alone, every backbone collapses substantially, but \textit{Our-MM} degrades the least and keeps the largest margin over \textit{MultiMAE} and the ViT-initialized baselines, suggesting the in-domain pretraining leaves the encoder less dependent on RGB acting as an anchor modality. Interestingly, \textit{MultiMAE} is the strongest model on the isolated Event stream for segmentation, a modality it never encountered during its own pretraining, which reinforces the robustness of its generic multimodal prior observed earlier in the paper. As soon as at least two modalities are available, all pretrained backbones recover sharply, and \textit{Our-MM} is consistently the best or second-best configuration across nearly every column; notably, the MAE-based models (\textit{MultiMAE} and \textit{Our-MM}) recover results much closer to their full-sensor performance than the other pretraining backbones, confirming that the gains from in-domain pretraining reported in Section~\ref{sec:evaluation} persist under degraded sensor availability. Comparing these numbers against Table~\ref{tab:dsec_main}, a checkpoint trained with all modalities and evaluated with missing ones still falls noticeably short of a model explicitly finetuned for that exact modality combination, indicating that dedicated per-configuration finetuning retains a meaningful edge over relying on one model exposed to all sensors at train time but only a subset at inference; closing this gap can be a promising direction for future work.

\section{Model scale}
\label{sec:model_scale}
Often the target devices where multimodal sensor fusion is performed are constrained in terms of computations, like in robotics and autonomous driving applications. Therefore, we further examine how prediction quality degrades with different pretraining sizes, comparing the ViT initialized variants of \textit{Base}, \textit{Small}, and \textit{Tiny} on SeeThroughFog (Table~\ref{tab:stf_scale}), where we can observe the multimodal degradation under different weather conditions. As expected, performance decreases largely monotonically with reduced capacity: \textit{Base} consistently leads across nearly all modality configurations and weather conditions, with Tiny trailing by a modest but consistent margin (typically 2-5 points), rather than collapsing outright. This suggests that pretraining based on the MAE objective transfers its multimodal alignment properties even to lightweight backbones, making the \textit{Small} and \textit{Tiny} variants viable candidates for deployment on compute-constrained embedded platforms. Interestingly, this ordering is not entirely strict under harsher conditions: for RGB-only and RGB+NIR inputs, \textit{Small} occasionally edges out \textit{Base} under fog (e.g. 63.46 vs. 61.15 on RGB dense fog, and 71.79 vs. 69.71 on RGB+NIR dense fog), hinting that smaller capacity may act as a mild regularizer under distribution shift, though this effect does not persist once additional physical sensors such as Depth or LiDAR are introduced.
As expected, all three scales exhibit the same qualitative weather degradation pattern independent of model size or modality: performance is highest under clear conditions, drops most sharply under dense fog, and snow/rain sits consistently between light and dense fog rather than at either extreme, mirroring the trend already observed in the main results (Table~\ref{tab:stf_main}). The consistency may indicate that robustness to weather conditions is more dependent on the pretraining strategy than backbone capacity, since even the \textit{Tiny} variant preserves the same relative gains across weather conditions as \textit{Base}.

\begin{table}[t]
\centering
\renewcommand{\arraystretch}{0.92}
\caption{%
  \textbf{Model Scale Comparison on STF.} We report 2D bounding box AP$_{R40}$ (\%) using the standard KITTI evaluation protocol
  for our ViT model at \textit{Base}, \textit{Small}, and \textit{Tiny} scales.
  The reported scores represent the average across all object classes for the easy, moderate, and hard difficulty levels.
  C\,=\,Clear; LF\,=\,Light Fog; DF\,=\,Dense Fog; S\,=\,Snow/Rain.
}
\label{tab:stf_scale}
\scriptsize
\setlength{\tabcolsep}{0pt}
\providecommand{\dcw}{}
\ifdefined\dcwlen\else\newlength{\dcwlen}\fi
\setlength{\dcwlen}{0.1025\columnwidth}
\begin{tabular}{@{}>{\raggedleft\arraybackslash}p{0.12\columnwidth}
  @{\ \ }|@{\ \ }
  >{\centering\arraybackslash}p{\dcwlen}
  @{}>{\centering\arraybackslash}p{\dcwlen}
  @{}>{\centering\arraybackslash}p{\dcwlen}
  @{}>{\centering\arraybackslash}p{\dcwlen}
  @{\ \ }|@{\ \ }
  >{\centering\arraybackslash}p{\dcwlen}
  @{}>{\centering\arraybackslash}p{\dcwlen}
  @{}>{\centering\arraybackslash}p{\dcwlen}
  @{}>{\centering\arraybackslash}p{\dcwlen}@{}}
\specialrule{.10em}{0pt}{0pt}
\specialrule{.04em}{2pt}{2pt}
& \textbf{C} & \textbf{LF} & \textbf{DF} & \textbf{S}
& \textbf{C} & \textbf{LF} & \textbf{DF} & \textbf{S} \\
\specialrule{.04em}{2pt}{0pt}
\noalign{\smallskip}
\multicolumn{1}{@{}p{0.12\columnwidth}@{}}{} & \multicolumn{4}{c}{\textit{RGB}}
& \multicolumn{4}{c}{\textit{NIR}} \\
\noalign{\smallskip}
\specialrule{.04em}{0pt}{2pt}
\textit{Base}  & 69.43 & 68.94 & 61.15 & 65.54 & 70.81 & 70.52 & 65.12 & 68.42 \\
\textit{Small} & 66.68 & 69.31 & 63.46 & 65.17 & 69.10 & 67.15 & 63.53 & 67.46 \\
\textit{Tiny}  & 65.22 & 65.53 & 60.63 & 62.59 & 67.00 & 68.39 & 58.66 & 65.35 \\
\specialrule{.04em}{2pt}{0pt}
\noalign{\smallskip}
\multicolumn{1}{@{}p{0.12\columnwidth}@{}}{} & \multicolumn{4}{c}{\textit{Depth}}
& \multicolumn{4}{c}{\textit{RGB+NIR}} \\
\noalign{\smallskip}
\specialrule{.04em}{0pt}{2pt}
\textit{Base}  & 46.68 & 41.48 & 19.60 & 40.98 & 75.08 & 76.77 & 69.71 & 73.92 \\
\textit{Small} & 44.00 & 39.71 & 14.59 & 37.27 & 74.25 & 75.22 & 71.79 & 72.12 \\
\textit{Tiny}  & 42.62 & 40.00 & 14.80 & 35.19 & 71.83 & 75.00 & 66.26 & 70.96 \\
\specialrule{.04em}{2pt}{0pt}
\noalign{\smallskip}
\multicolumn{1}{@{}p{0.12\columnwidth}@{}}{} & \multicolumn{4}{c}{\textit{RGB+Depth}}
& \multicolumn{4}{c}{\textit{RGB+LiDAR}} \\
\noalign{\smallskip}
\specialrule{.04em}{0pt}{2pt}
\textit{Base}  & 70.94 & 70.59 & 58.78 & 67.14 & 70.02 & 70.35 & 64.42 & 66.94 \\
\textit{Small} & 68.36 & 68.69 & 53.90 & 65.09 & 67.64 & 68.25 & 61.51 & 65.98 \\
\textit{Tiny}  & 65.71 & 67.68 & 51.04 & 62.86 & 65.34 & 66.32 & 56.91 & 62.50 \\
\specialrule{.04em}{2pt}{0pt}
\noalign{\smallskip}
\multicolumn{1}{@{}p{0.12\columnwidth}@{}}{} & \multicolumn{4}{c}{\textit{RGB+NIR+Depth}}
& \multicolumn{4}{c}{\textit{RGB+NIR+LiDAR}} \\
\noalign{\smallskip}
\specialrule{.04em}{0pt}{2pt}
\textit{Base}  & 75.27 & 78.92 & 67.15 & 73.34 & 75.17 & 78.06 & 69.75 & 74.07 \\
\textit{Small} & 74.61 & 77.38 & 63.08 & 72.49 & 74.04 & 75.54 & 69.19 & 72.56 \\
\textit{Tiny}  & 71.44 & 75.57 & 59.02 & 70.37 & 72.94 & 75.72 & 65.52 & 71.60 \\
\specialrule{.04em}{2pt}{2pt}
\specialrule{.10em}{0pt}{0pt}
\end{tabular}
\end{table}

%% file: appendix/licensing.tex
\section{Datasets licensing and reproducibility}
\label{sec:licensing}
Our pretraining corpus aggregates 16 datasets under heterogeneous licenses, several of which (Cityscapes\cite{Cordts2016Cityscapes}, LLVIP\cite{jia2021llvip}, Freiburg Thermal\cite{vertens2020heatnet}) explicitly prohibit redistribution of the data or derivatives, and others (KAIST-MS-Pedestrian\cite{7298706}, MS2\cite{shin2023deep}, RASMD\cite{JIN2026103872}) are released under CC BY-NC-SA terms whose share-alike clause is incompatible with unifying them under a single collection license. Several remaining constituent datasets (ECMD\cite{10342726}, FMB\cite{liu2023segmif}, MFNet\cite{8206396}, NTU4DRadLM\cite{ZhangZhugeLiu2023ITSC}, SMOD\cite{11146667}, STheReO\cite{9981857}, Vivid++\cite{9760091}) provide no explicit redistribution license at all. We therefore cannot legally redistribute the aggregated corpus. Table~\ref{tab:dataset_licenses} summarizes the licensing terms of each constituent dataset. To support reproducibility, we release the weights pretrained on this corpus, the script to reproduce the data collection, as well as the full evaluation framework to reproduce the presented results, at \href{https://github.com/DTU-PAS/sensormae}{\textit{https://github.com/DTU-PAS/sensormae}}. The construction of the pretraining collection itself relies on a mix of automatic and manual steps, which we detail in Appendix \ref{sec:datasets}. This should allow the independent reconstruction of an equivalent corpus. The pretraining codebase is heavily based on the MultiMAE\cite{multimae} repository\footnote{\url{https://github.com/EPFL-VILAB/MultiMAE}}, with the targeted modifications described in Section~\ref{sec:methods}.

\begin{table}[t]
\centering
\caption{Licensing terms of the 16 constituent datasets in our
pretraining collection. Of the datasets marked ``Not specified'' we could not find a published license or terms-of-use document accompanying the public release.}
\label{tab:dataset_licenses}
\resizebox{\columnwidth}{!}{%
\begin{tabular}{@{}ll@{}}
\toprule
\textbf{Dataset} & \textbf{License / Terms} \\
\midrule
BDD100K\cite{Yu2018BDD100KAD} & UC Berkeley custom (BDD100K License) \\
Cityscapes\cite{Cordts2016Cityscapes} & Custom non-commercial license \\
DSEC\cite{gehrig2021dsec, Gehrig3dv2021} & CC BY-SA 4.0 \\
ECMD\cite{10342726} & Not specified \\
FMB\cite{liu2023segmif} & Not specified \\
Freiburg Thermal\cite{vertens2020heatnet} & Custom (non-commercial) \\
KAIST-MS-Pedestrian\cite{7298706} & CC BY-NC-SA 3.0 \\
LLVIP\cite{jia2021llvip} & Custom non-commercial license \\
MFNet\cite{8206396} & Not specified \\
MS2\cite{shin2023deep} & CC BY-NC-SA 3.0 \\
NSAVP\cite{carmichael2025dataset} & CC BY 4.0 (Deep Blue Data default) \\
NTU4DRadLM\cite{ZhangZhugeLiu2023ITSC} & Not specified \\
RASMD\cite{JIN2026103872} & CC BY-NC-SA 4.0 \\
SMOD\cite{11146667} & Not specified \\
STheReO\cite{9981857} & Not specified \\
Vivid++\cite{9760091} & Not specified \\
\bottomrule
\end{tabular}%
}
\end{table}

%% file: appendix/samples.tex
\begin{figure*}[b]
\centering

\begin{minipage}{0.48\linewidth}
\includegraphics[width=\linewidth]{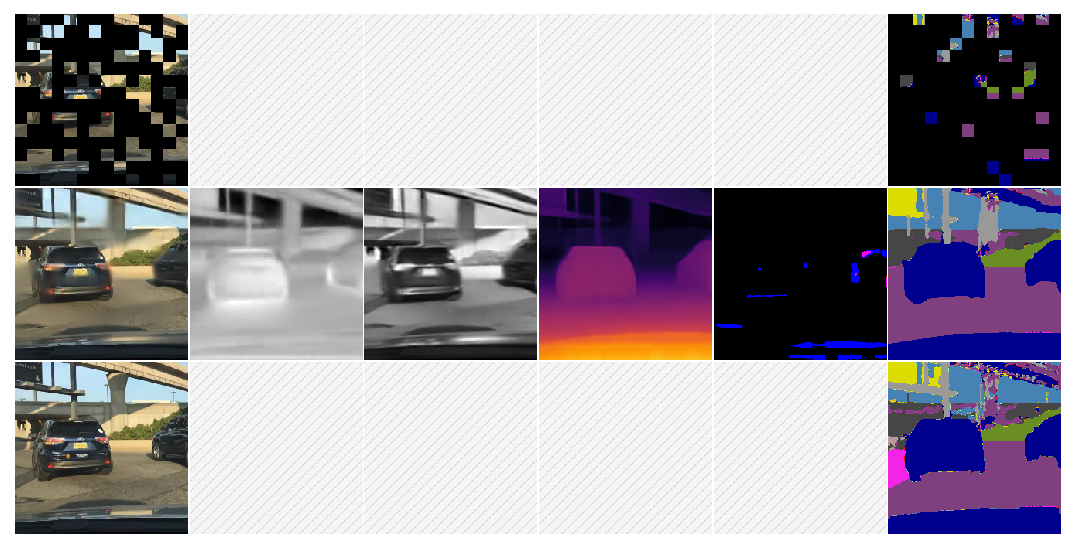}\\[0.05cm]
\includegraphics[width=\linewidth]{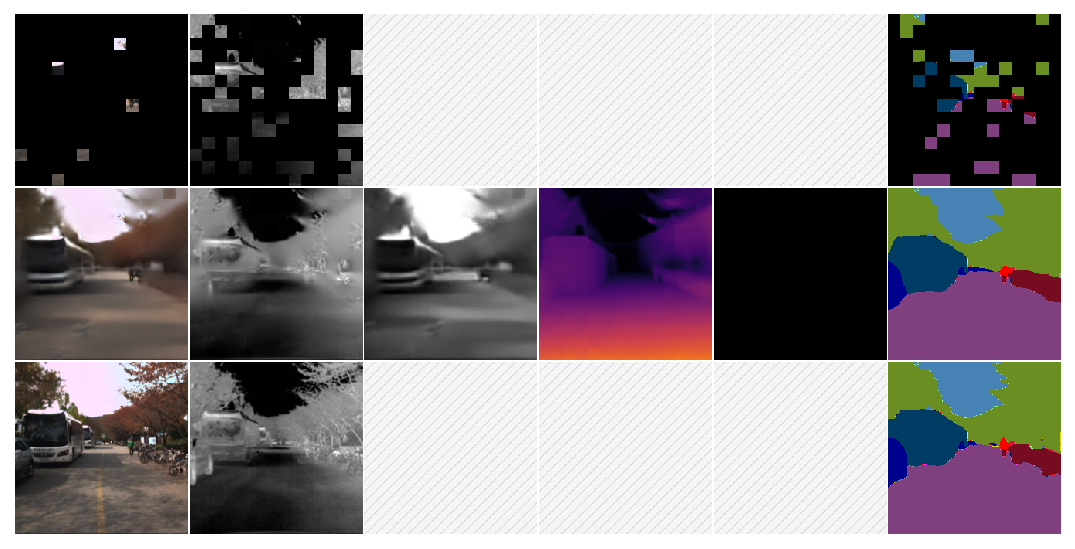}\\[0.05cm]
\includegraphics[width=\linewidth]{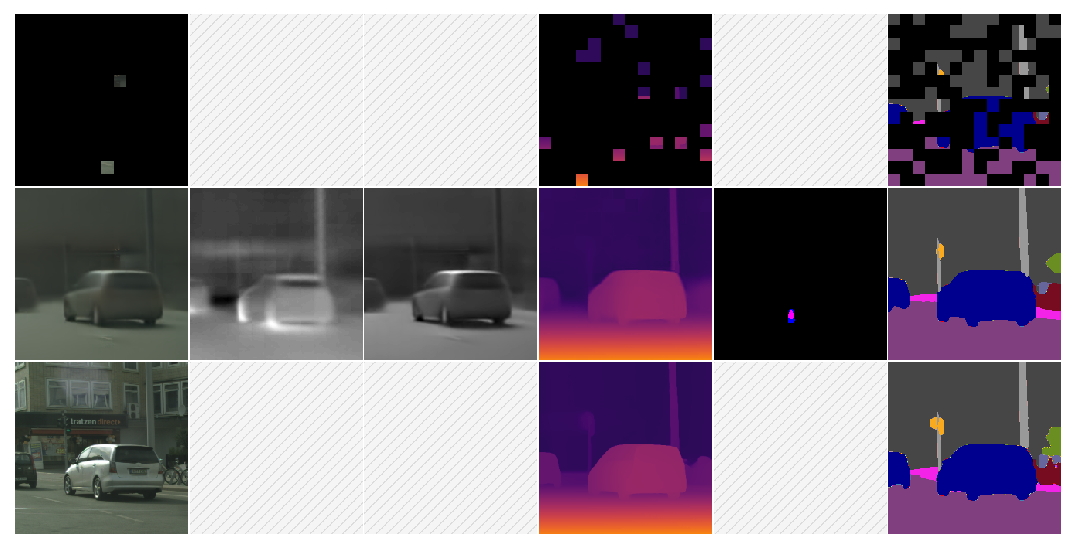}\\[0.05cm]
\includegraphics[width=\linewidth]{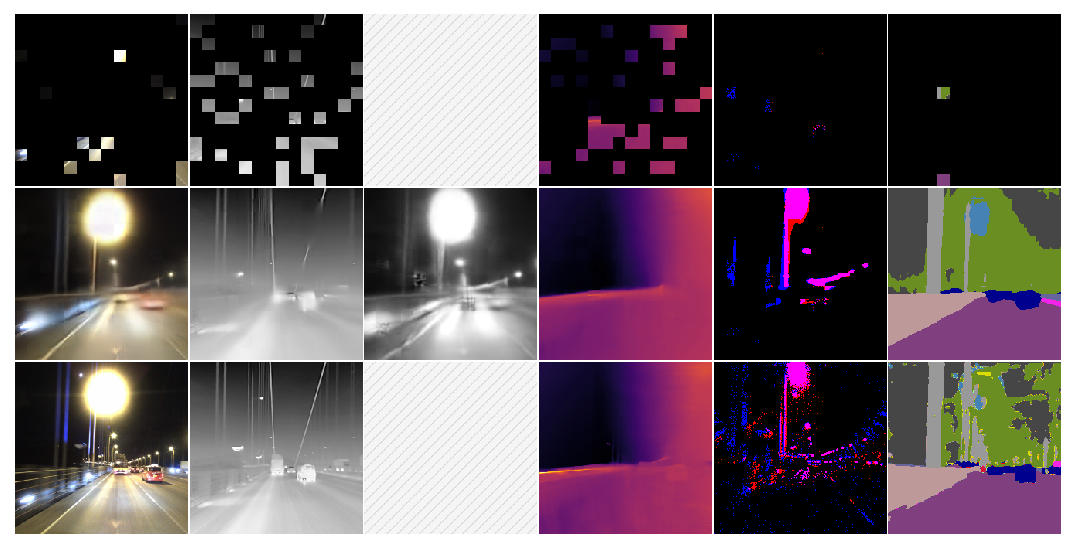}\\[0.05cm]
\includegraphics[width=\linewidth]{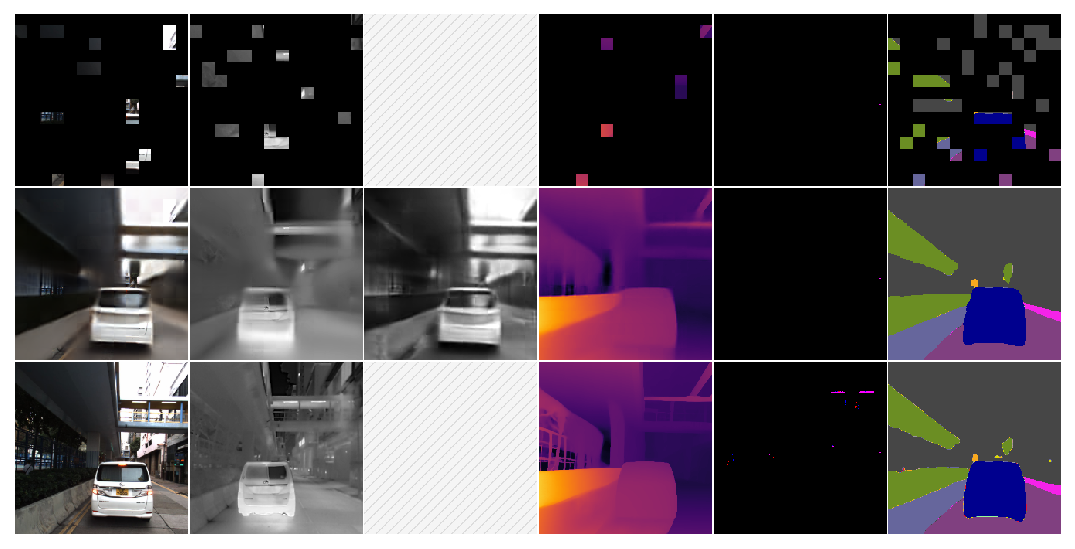}
\end{minipage}
\hfill
\begin{minipage}{0.48\linewidth}
\includegraphics[width=\linewidth]{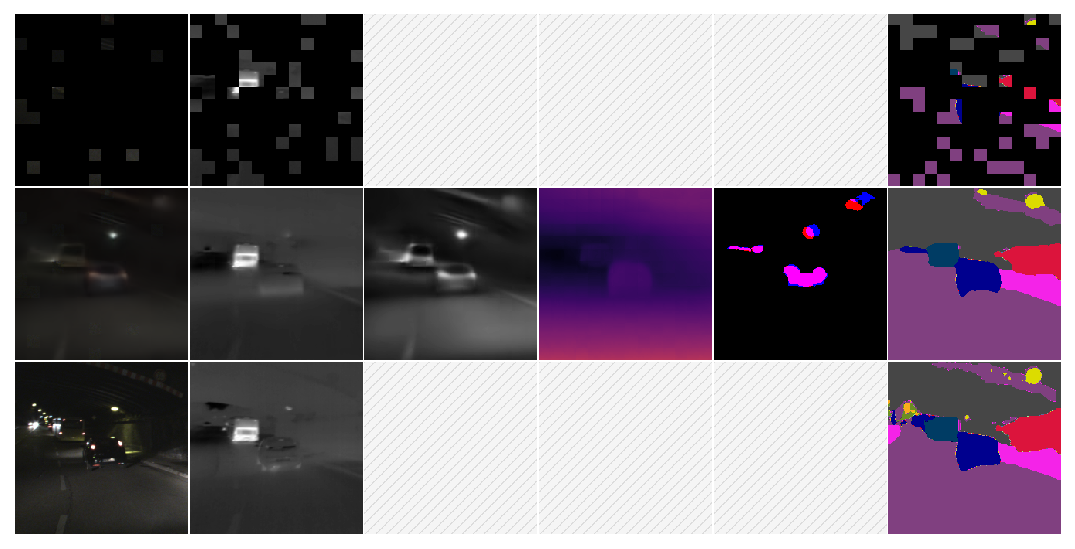}\\[0.05cm]
\includegraphics[width=\linewidth]{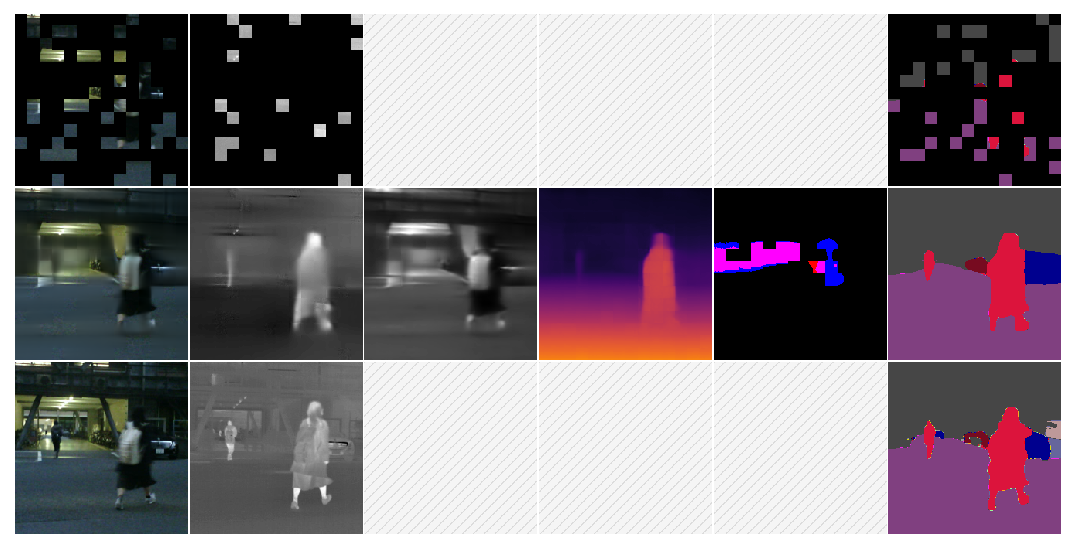}\\[0.05cm]
\includegraphics[width=\linewidth]{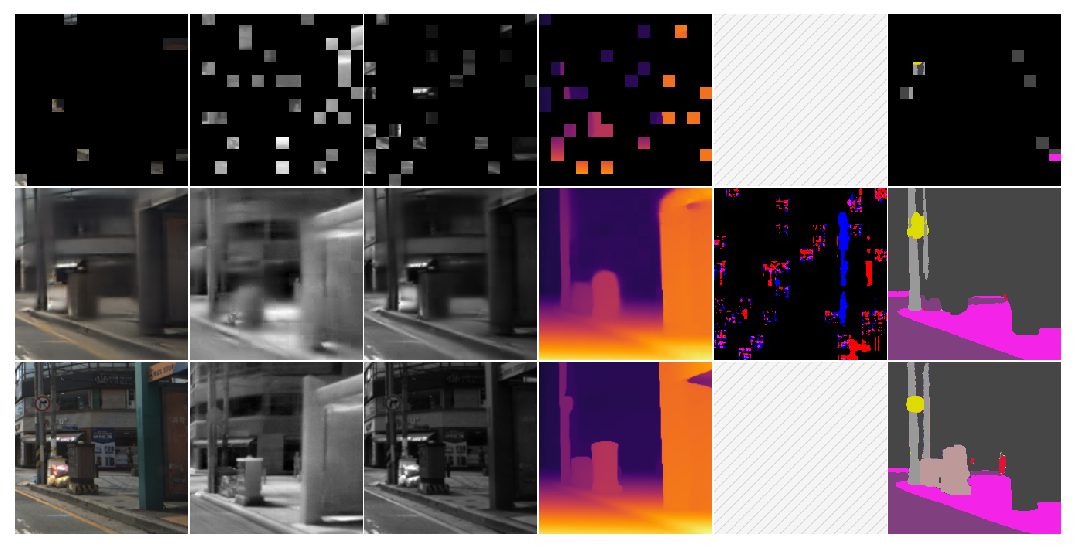}\\[0.05cm]
\includegraphics[width=\linewidth]{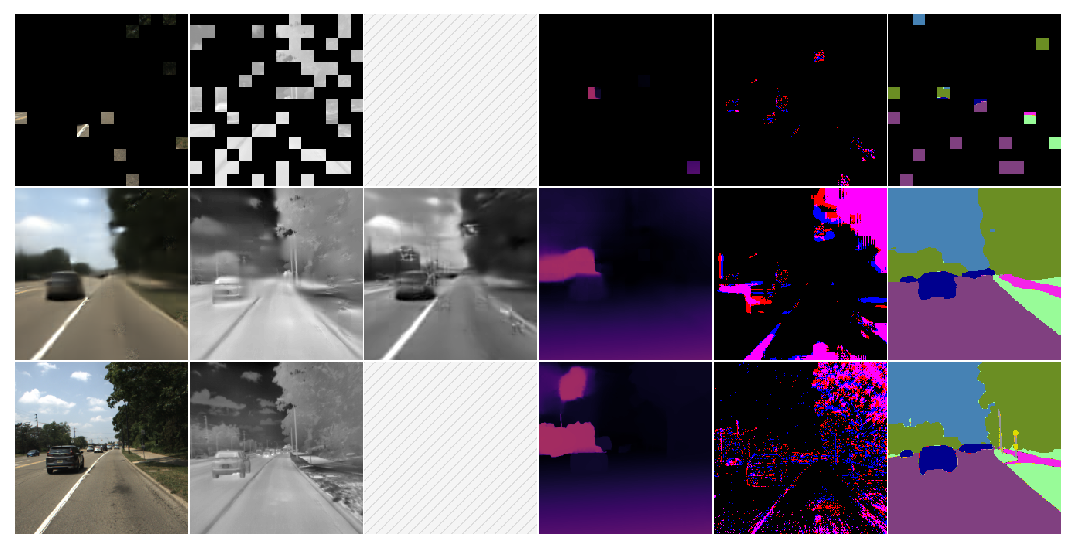}\\[0.05cm]
\includegraphics[width=\linewidth]{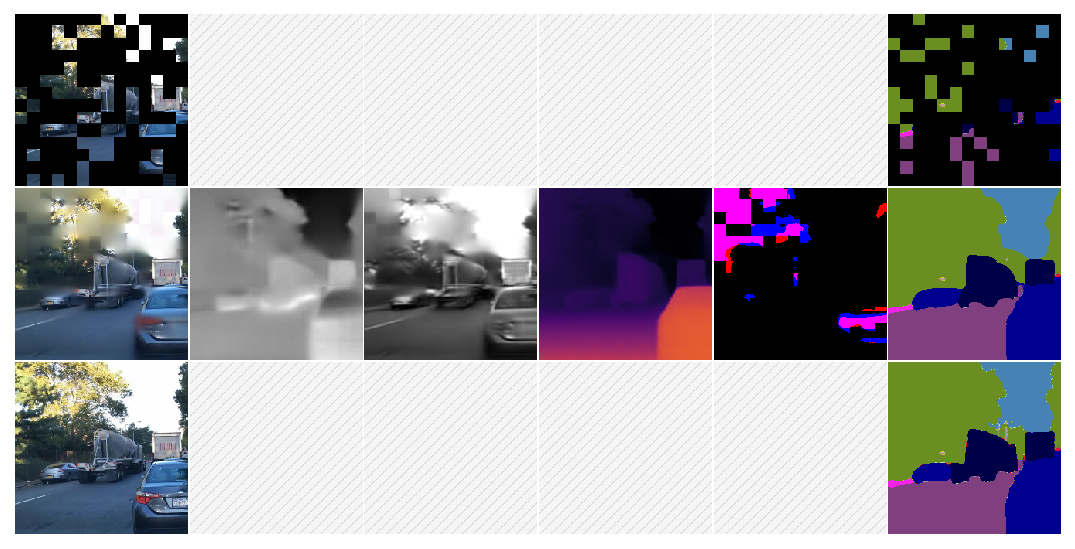}
\end{minipage}

\caption{Extended qualitative cross-modal reconstruction results across ten distinct driving scenarios from the pretraining corpus. The first row represents the input, second row shows the model output, while the last row shows the target. For each image, from left to right: RGB, NIR, LWIR, Depth, Event, Segmentation. Grey images represent unavailable input/target modalities.}
\label{fig:appendix_banners_combined}
\end{figure*}

\begin{figure*}[t]
    \centering
    \includegraphics[width=1.0\linewidth]{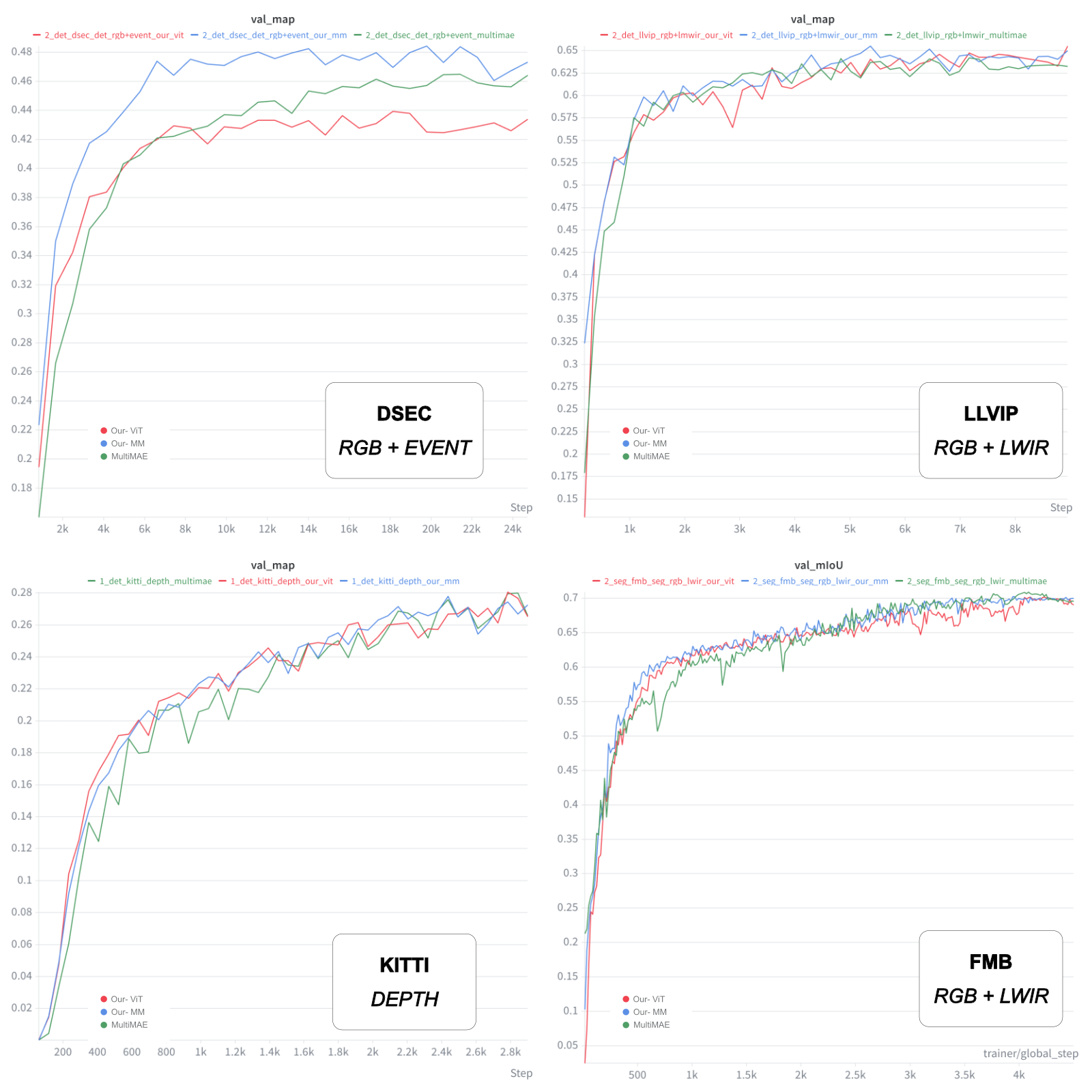}
\caption{
    \textbf{Convergence analysis across datasets.} Comparison of learning curves between \textit{Our-MM} (blue), \textit{Our-ViT} (red) and \textit{MultiMAE} (green) across different benchmarks: DSEC, LLVIP, KITTI for Object Detection and FMB for Semantic Segmentation. Our in-domain pretrained weights present steeper learning curves at early training steps, providing a stronger starting point for adapting to new sensor modalities. The performance gap narrows as training progresses, which suggests that once \textit{MultiMAE} catches up to the target modality, it is able to use its strong cross attention capabilities.
}
\label{fig:learn_curves}
\end{figure*}

%% file: main.bbl
\begin{thebibliography}{43}
\providecommand{\natexlab}[1]{#1}
\providecommand{\url}[1]{\texttt{#1}}
\expandafter\ifx\csname urlstyle\endcsname\relax
  \providecommand{\doi}[1]{doi: #1}\else
  \providecommand{\doi}{doi: \begingroup \urlstyle{rm}\Url}\fi

\bibitem[Bachmann et~al.(2022)Bachmann, Mizrahi, Atanov, and Zamir]{multimae}
Roman Bachmann, David Mizrahi, Andrei Atanov, and Amir Zamir.
\newblock {MultiMAE}: Multi-modal multi-task masked autoencoders.
\newblock In \emph{Proceedings of the European Conference on Computer Vision
  (ECCV)}, pages 348--367. Springer, 2022.

\bibitem[Bachmann et~al.(2024)Bachmann, Kar, Mizrahi, Garjani, Gao, Griffiths,
  Hu, Dehghan, and Zamir]{2024_4m21}
Roman Bachmann, O{\u{g}}uzhan~Fatih Kar, David Mizrahi, Ali Garjani, Mingfei
  Gao, David Griffiths, Jiaming Hu, Afshin Dehghan, and Amir Zamir.
\newblock {4M-21}: An any-to-any vision model for tens of tasks and modalities.
\newblock In \emph{Advances in Neural Information Processing Systems
  (NeurIPS)}, pages 61872--61911, 2024.

\bibitem[Bijelic et~al.(2020)Bijelic, Gruber, Mannan, Kraus, Ritter, Dietmayer,
  and Heide]{bijelic2020seeing}
Mario Bijelic, Tobias Gruber, Fahim Mannan, Florian Kraus, Werner Ritter, Klaus
  Dietmayer, and Felix Heide.
\newblock Seeing through fog without seeing fog: Deep multimodal sensor fusion
  in unseen adverse weather.
\newblock In \emph{Proceedings of the IEEE/CVF Conference on Computer Vision
  and Pattern Recognition (CVPR)}, pages 11682--11692, 2020.

\bibitem[Cai and Vasconcelos(2018)]{cai2018cascadercnn}
Zhaowei Cai and Nuno Vasconcelos.
\newblock Cascade {R-CNN}: Delving into high quality object detection.
\newblock In \emph{Proceedings of the IEEE Conference on Computer Vision and
  Pattern Recognition (CVPR)}, pages 6154--6162, 2018.

\bibitem[Carlesso et~al.(2026)Carlesso, Mothe, and
  Ionescu]{carlesso2025curriculum}
Hugo Carlesso, Josiane Mothe, and Radu~Tudor Ionescu.
\newblock Curriculum multi-task self-supervision improves lightweight
  architectures for onboard satellite hyperspectral image segmentation.
\newblock In \emph{Proceedings of the IEEE International Conference on Robotics
  and Automation (ICRA)}, pages 2180--2187, 2026.

\bibitem[Carmichael et~al.(2025)Carmichael, Buchan, Ramanagopal, Ravi,
  Vasudevan, and Skinner]{carmichael2025dataset}
Spencer Carmichael, Austin Buchan, Mani Ramanagopal, Radhika Ravi, Ram
  Vasudevan, and Katherine~A. Skinner.
\newblock Dataset and benchmark: Novel sensors for autonomous vehicle
  perception.
\newblock \emph{The International Journal of Robotics Research}, 44\penalty0
  (3):\penalty0 355--365, 2025.

\bibitem[Caron et~al.(2021)Caron, Touvron, Misra, J{\'e}gou, Mairal,
  Bojanowski, and Joulin]{dino}
Mathilde Caron, Hugo Touvron, Ishan Misra, Herv{\'e} J{\'e}gou, Julien Mairal,
  Piotr Bojanowski, and Armand Joulin.
\newblock Emerging properties in self-supervised vision transformers.
\newblock In \emph{Proceedings of the IEEE/CVF International Conference on
  Computer Vision (ICCV)}, pages 9650--9660, 2021.

\bibitem[Chen et~al.(2024)Chen, Guan, Huang, Zhong, Wen, Hsu, and Lu]{10342726}
Peiyu Chen, Weipeng Guan, Feng Huang, Yihan Zhong, Weisong Wen, Li-Ta Hsu, and
  Peng Lu.
\newblock {ECMD}: An event-centric multisensory driving dataset for {SLAM}.
\newblock \emph{IEEE Transactions on Intelligent Vehicles}, 9\penalty0
  (1):\penalty0 407--416, 2024.

\bibitem[Chen et~al.(2020)Chen, Kornblith, Norouzi, and Hinton]{SIMCLR}
Ting Chen, Simon Kornblith, Mohammad Norouzi, and Geoffrey Hinton.
\newblock A simple framework for contrastive learning of visual
  representations.
\newblock In \emph{Proceedings of the International Conference on Machine
  Learning (ICML)}, pages 1597--1607. PMLR, 2020.

\bibitem[Chen et~al.(2025)Chen, Qian, Yang, Wang, and Yang]{11146667}
Zizhao Chen, Yeqiang Qian, Xiaoxiao Yang, Chunxiang Wang, and Ming Yang.
\newblock {AMFD}: Distillation via adaptive multimodal fusion for multispectral
  pedestrian detection.
\newblock \emph{IEEE Transactions on Multimedia}, 27:\penalty0 8298--8310,
  2025.

\bibitem[Cheng et~al.(2022)Cheng, Misra, Schwing, Kirillov, and
  Girdhar]{cheng2022masked}
Bowen Cheng, Ishan Misra, Alexander~G. Schwing, Alexander Kirillov, and Rohit
  Girdhar.
\newblock Masked-attention mask transformer for universal image segmentation.
\newblock In \emph{Proceedings of the IEEE/CVF Conference on Computer Vision
  and Pattern Recognition (CVPR)}, pages 1290--1299, 2022.

\bibitem[Cordts et~al.(2016)Cordts, Omran, Ramos, Rehfeld, Enzweiler, Benenson,
  Franke, Roth, and Schiele]{Cordts2016Cityscapes}
Marius Cordts, Mohamed Omran, Sebastian Ramos, Timo Rehfeld, Markus Enzweiler,
  Rodrigo Benenson, Uwe Franke, Stefan Roth, and Bernt Schiele.
\newblock The {Cityscapes} dataset for semantic urban scene understanding.
\newblock In \emph{Proceedings of the IEEE Conference on Computer Vision and
  Pattern Recognition (CVPR)}, pages 3213--3223, 2016.

\bibitem[Deng et~al.(2009)Deng, Dong, Socher, Li, Li, and
  Fei-Fei]{deng2009imagenet}
Jia Deng, Wei Dong, Richard Socher, Li-Jia Li, Kai Li, and Li Fei-Fei.
\newblock {ImageNet}: A large-scale hierarchical image database.
\newblock In \emph{Proceedings of the IEEE Conference on Computer Vision and
  Pattern Recognition (CVPR)}, pages 248--255, 2009.

\bibitem[Dosovitskiy et~al.(2021)Dosovitskiy, Beyer, Kolesnikov, Weissenborn,
  Zhai, Unterthiner, Dehghani, Minderer, Heigold, Gelly, Uszkoreit, and
  Houlsby]{dosovitskiy2020image}
Alexey Dosovitskiy, Lucas Beyer, Alexander Kolesnikov, Dirk Weissenborn,
  Xiaohua Zhai, Thomas Unterthiner, Mostafa Dehghani, Matthias Minderer, Georg
  Heigold, Sylvain Gelly, Jakob Uszkoreit, and Neil Houlsby.
\newblock An image is worth 16x16 words: Transformers for image recognition at
  scale.
\newblock In \emph{International Conference on Learning Representations
  (ICLR)}, 2021.

\bibitem[Gehrig and Scaramuzza(2024)]{gehrig2024low}
Daniel Gehrig and Davide Scaramuzza.
\newblock Low-latency automotive vision with event cameras.
\newblock \emph{Nature}, 629\penalty0 (8014):\penalty0 1034--1040, 2024.

\bibitem[Gehrig et~al.(2021{\natexlab{a}})Gehrig, Aarents, Gehrig, and
  Scaramuzza]{gehrig2021dsec}
Mathias Gehrig, Willem Aarents, Daniel Gehrig, and Davide Scaramuzza.
\newblock {DSEC}: A stereo event camera dataset for driving scenarios.
\newblock \emph{IEEE Robotics and Automation Letters}, 6\penalty0 (3):\penalty0
  4947--4954, 2021{\natexlab{a}}.

\bibitem[Gehrig et~al.(2021{\natexlab{b}})Gehrig, Millh{\"a}usler, Gehrig, and
  Scaramuzza]{Gehrig3dv2021}
Mathias Gehrig, Mario Millh{\"a}usler, Daniel Gehrig, and Davide Scaramuzza.
\newblock {E-RAFT}: Dense optical flow from event cameras.
\newblock In \emph{International Conference on 3D Vision (3DV)}, pages
  197--206, 2021{\natexlab{b}}.

\bibitem[Geiger et~al.(2012)Geiger, Lenz, and Urtasun]{Geiger2012CVPR}
Andreas Geiger, Philip Lenz, and Raquel Urtasun.
\newblock Are we ready for autonomous driving? {The} {KITTI} vision benchmark
  suite.
\newblock In \emph{Proceedings of the IEEE Conference on Computer Vision and
  Pattern Recognition (CVPR)}, pages 3354--3361, 2012.

\bibitem[Ha et~al.(2017)Ha, Watanabe, Karasawa, Ushiku, and Harada]{8206396}
Qishen Ha, Kohei Watanabe, Takumi Karasawa, Yoshitaka Ushiku, and Tatsuya
  Harada.
\newblock {MFNet}: Towards real-time semantic segmentation for autonomous
  vehicles with multi-spectral scenes.
\newblock In \emph{Proceedings of the IEEE/RSJ International Conference on
  Intelligent Robots and Systems (IROS)}, pages 5108--5115, 2017.

\bibitem[He et~al.(2020)He, Fan, Wu, Xie, and Girshick]{he2019momentum}
Kaiming He, Haoqi Fan, Yuxin Wu, Saining Xie, and Ross Girshick.
\newblock Momentum contrast for unsupervised visual representation learning.
\newblock In \emph{Proceedings of the IEEE/CVF Conference on Computer Vision
  and Pattern Recognition (CVPR)}, pages 9726--9735, 2020.

\bibitem[He et~al.(2022)He, Chen, Xie, Li, Doll{\'a}r, and
  Girshick]{he2022masked}
Kaiming He, Xinlei Chen, Saining Xie, Yanghao Li, Piotr Doll{\'a}r, and Ross
  Girshick.
\newblock Masked autoencoders are scalable vision learners.
\newblock In \emph{Proceedings of the IEEE/CVF Conference on Computer Vision
  and Pattern Recognition (CVPR)}, pages 16000--16009, 2022.

\bibitem[Hwang et~al.(2015)Hwang, Park, Kim, Choi, and Kweon]{7298706}
Soonmin Hwang, Jaesik Park, Namil Kim, Yukyung Choi, and In~So Kweon.
\newblock Multispectral pedestrian detection: Benchmark dataset and baseline.
\newblock In \emph{Proceedings of the IEEE Conference on Computer Vision and
  Pattern Recognition (CVPR)}, pages 1037--1045, 2015.

\bibitem[Jia et~al.(2021)Jia, Zhu, Li, Tang, and Zhou]{jia2021llvip}
Xinyu Jia, Chuang Zhu, Minzhen Li, Wenqi Tang, and Wenli Zhou.
\newblock {LLVIP}: A visible-infrared paired dataset for low-light vision.
\newblock In \emph{Proceedings of the IEEE/CVF International Conference on
  Computer Vision Workshops (ICCVW)}, pages 3496--3504, 2021.

\bibitem[Jin et~al.(2026)Jin, Kovac, Nalcakan, Park, Yeo, Ju, and
  Kim]{JIN2026103872}
Youngwan Jin, Michal Kovac, Yagiz Nalcakan, Incheol Park, Sanghyeop Yeo,
  Hyeongjin Ju, and Shiho Kim.
\newblock {RASMD}: {RGB} and {SWIR} multispectral driving dataset for robust
  perception in adverse conditions.
\newblock \emph{Information Fusion}, 128:\penalty0 103872, 2026.

\bibitem[Kerbl et~al.(2023)Kerbl, Kopanas, Leimk{\"u}hler, and
  Drettakis]{kerbl20233d}
Bernhard Kerbl, Georgios Kopanas, Thomas Leimk{\"u}hler, and George Drettakis.
\newblock {3D} {Gaussian} splatting for real-time radiance field rendering.
\newblock \emph{ACM Transactions on Graphics}, 42\penalty0 (4):\penalty0
  139:1--139:14, 2023.

\bibitem[Kraus et~al.(2024)Kraus, Kenyon-Dean, Saberian, Fallah, McLean, Leung,
  Sharma, Khan, Balakrishnan, Celik, et~al.]{kraus2024masked}
Oren Kraus, Kian Kenyon-Dean, Saber Saberian, Maryam Fallah, Peter McLean, Jess
  Leung, Vasudev Sharma, Ayla Khan, Jia Balakrishnan, Safiye Celik, et~al.
\newblock Masked autoencoders for microscopy are scalable learners of cellular
  biology.
\newblock In \emph{Proceedings of the IEEE/CVF Conference on Computer Vision
  and Pattern Recognition (CVPR)}, pages 11757--11768, 2024.

\bibitem[Lee et~al.(2022)Lee, Cho, Shin, Kim, and Myung]{9760091}
Alex~Junho Lee, Younggun Cho, Young-sik Shin, Ayoung Kim, and Hyun Myung.
\newblock {ViViD++}: Vision for visibility dataset.
\newblock \emph{IEEE Robotics and Automation Letters}, 7\penalty0 (3):\penalty0
  6282--6289, 2022.

\bibitem[Liu et~al.(2023)Liu, Liu, Wu, Ma, Liu, Zhong, Luo, and
  Fan]{liu2023segmif}
Jinyuan Liu, Zhu Liu, Guanyao Wu, Long Ma, Risheng Liu, Wei Zhong, Zhongxuan
  Luo, and Xin Fan.
\newblock Multi-interactive feature learning and a full-time multi-modality
  benchmark for image fusion and segmentation.
\newblock In \emph{Proceedings of the IEEE/CVF International Conference on
  Computer Vision (ICCV)}, pages 8115--8124, 2023.

\bibitem[Mildenhall et~al.(2021)Mildenhall, Srinivasan, Tancik, Barron,
  Ramamoorthi, and Ng]{mildenhall2021nerf}
Ben Mildenhall, Pratul~P. Srinivasan, Matthew Tancik, Jonathan~T. Barron, Ravi
  Ramamoorthi, and Ren Ng.
\newblock {NeRF}: Representing scenes as neural radiance fields for view
  synthesis.
\newblock \emph{Communications of the ACM}, 65\penalty0 (1):\penalty0 99--106,
  2021.

\bibitem[Mizrahi et~al.(2023)Mizrahi, Bachmann, Kar, Yeo, Gao, Dehghan, and
  Zamir]{2023_4m}
David Mizrahi, Roman Bachmann, O{\u{g}}uzhan~Fatih Kar, Teresa Yeo, Mingfei
  Gao, Afshin Dehghan, and Amir Zamir.
\newblock {4M}: Massively multimodal masked modeling.
\newblock In \emph{Advances in Neural Information Processing Systems
  (NeurIPS)}, pages 58363--58408, 2023.

\bibitem[Pham et~al.(2025)Pham, Caicedo, and Plummer]{PhamChaMAE2025}
Chau Pham, Juan~C. Caicedo, and Bryan~A. Plummer.
\newblock {ChA-MAEViT}: Unifying channel-aware masked autoencoders and
  multi-channel vision transformers for improved cross-channel learning.
\newblock In \emph{Advances in Neural Information Processing Systems
  (NeurIPS)}, pages 13342--13372, 2025.

\bibitem[Pizer et~al.(1990)Pizer, Johnston, Ericksen, Yankaskas, and
  Muller]{109340}
Stephen~M. Pizer, R.~Eugene Johnston, James~P. Ericksen, Bonnie~C. Yankaskas,
  and Keith~E. Muller.
\newblock Contrast-limited adaptive histogram equalization: speed and
  effectiveness.
\newblock In \emph{Proceedings of the First Conference on Visualization in
  Biomedical Computing}, pages 337--345, 1990.

\bibitem[Radford et~al.(2021)Radford, Kim, Hallacy, Ramesh, Goh, Agarwal,
  Sastry, Askell, Mishkin, Clark, Krueger, and Sutskever]{radford2021learning}
Alec Radford, Jong~Wook Kim, Chris Hallacy, Aditya Ramesh, Gabriel Goh,
  Sandhini Agarwal, Girish Sastry, Amanda Askell, Pamela Mishkin, Jack Clark,
  Gretchen Krueger, and Ilya Sutskever.
\newblock Learning transferable visual models from natural language
  supervision.
\newblock In \emph{Proceedings of the International Conference on Machine
  Learning (ICML)}, pages 8748--8763. PMLR, 2021.

\bibitem[Ranftl et~al.(2021)Ranftl, Bochkovskiy, and Koltun]{ranftl2021vision}
Ren{\'e} Ranftl, Alexey Bochkovskiy, and Vladlen Koltun.
\newblock Vision transformers for dense prediction.
\newblock In \emph{Proceedings of the IEEE/CVF International Conference on
  Computer Vision (ICCV)}, pages 12179--12188, 2021.

\bibitem[Shin et~al.(2023)Shin, Park, and Kweon]{shin2023deep}
Ukcheol Shin, Jinsun Park, and In~So Kweon.
\newblock Deep depth estimation from thermal image.
\newblock In \emph{Proceedings of the IEEE/CVF Conference on Computer Vision
  and Pattern Recognition (CVPR)}, pages 1043--1053, 2023.

\bibitem[Strudel et~al.(2021)Strudel, Garcia, Laptev, and
  Schmid]{strudel2021segmenter}
Robin Strudel, Ricardo Garcia, Ivan Laptev, and Cordelia Schmid.
\newblock Segmenter: Transformer for semantic segmentation.
\newblock In \emph{Proceedings of the IEEE/CVF International Conference on
  Computer Vision (ICCV)}, pages 7262--7272, 2021.

\bibitem[Sun et~al.(2022)Sun, Messikommer, Gehrig, and Scaramuzza]{sun2022ess}
Zhaoning Sun, Nico Messikommer, Daniel Gehrig, and Davide Scaramuzza.
\newblock {ESS}: Learning event-based semantic segmentation from still images.
\newblock In \emph{Proceedings of the European Conference on Computer Vision
  (ECCV)}, pages 341--357. Springer, 2022.

\bibitem[Vertens et~al.(2020)Vertens, Z{\"u}rn, and
  Burgard]{vertens2020heatnet}
Johan Vertens, Jannik Z{\"u}rn, and Wolfram Burgard.
\newblock {HeatNet}: Bridging the day-night domain gap in semantic segmentation
  with thermal images.
\newblock In \emph{Proceedings of the IEEE/RSJ International Conference on
  Intelligent Robots and Systems (IROS)}, pages 8461--8468, 2020.

\bibitem[Wang et~al.(2025)Wang, Hern{\'a}ndez~Hern{\'a}ndez, Albrecht, and
  Zhu]{wang2024feature}
Yi Wang, Hugo Hern{\'a}ndez~Hern{\'a}ndez, Conrad~M. Albrecht, and Xiao~Xiang
  Zhu.
\newblock Feature guided masked autoencoder for self-supervised learning in
  remote sensing.
\newblock \emph{IEEE Journal of Selected Topics in Applied Earth Observations
  and Remote Sensing}, 18:\penalty0 321--336, 2025.

\bibitem[Wen et~al.(2025)Wen, Trepte, Aribido, Kautz, Gallo, and
  Birchfield]{wen_trepte_aribido_kautz_gallo_birchfield}
Bowen Wen, Matthew Trepte, Joseph Aribido, Jan Kautz, Orazio Gallo, and Stan
  Birchfield.
\newblock {FoundationStereo}: Zero-shot stereo matching.
\newblock In \emph{Proceedings of the IEEE/CVF Conference on Computer Vision
  and Pattern Recognition (CVPR)}, pages 5249--5260, 2025.

\bibitem[Yu et~al.(2020)Yu, Chen, Wang, Xian, Chen, Liu, Madhavan, and
  Darrell]{Yu2018BDD100KAD}
Fisher Yu, Haofeng Chen, Xin Wang, Wenqi Xian, Yingying Chen, Fangchen Liu,
  Vashisht Madhavan, and Trevor Darrell.
\newblock {BDD100K}: A diverse driving dataset for heterogeneous multitask
  learning.
\newblock In \emph{Proceedings of the IEEE/CVF Conference on Computer Vision
  and Pattern Recognition (CVPR)}, pages 2633--2642, 2020.

\bibitem[Yun et~al.(2022)Yun, Jung, Kim, Jung, Cho, Jeon, Kim, and
  Kim]{9981857}
Seungsang Yun, Minwoo Jung, Jeongyun Kim, Sangwoo Jung, Younghun Cho,
  Myung-Hwan Jeon, Giseop Kim, and Ayoung Kim.
\newblock {STheReO}: Stereo thermal dataset for research in odometry and
  mapping.
\newblock In \emph{Proceedings of the IEEE/RSJ International Conference on
  Intelligent Robots and Systems (IROS)}, pages 3857--3864, 2022.

\bibitem[Zhang et~al.(2023)Zhang, Zhuge, Liu, Peng, Wu, Zhang, Lyu, Li, Zhao,
  Kircali, Mharolkar, Yang, Yi, Wang, and Wang]{ZhangZhugeLiu2023ITSC}
Jun Zhang, Huayang Zhuge, Yiyao Liu, Guohao Peng, Zhenyu Wu, Haoyuan Zhang,
  Qiyang Lyu, Heshan Li, Chunyang Zhao, Dogan Kircali, Sanat Mharolkar, Xun
  Yang, Su Yi, Yuanzhe Wang, and Danwei Wang.
\newblock {NTU4DRadLM}: {4D} radar-centric multi-modal dataset for localization
  and mapping.
\newblock In \emph{Proceedings of the IEEE International Conference on
  Intelligent Transportation Systems (ITSC)}, pages 4291--4296, 2023.

\end{thebibliography}
